\documentclass[10pt,twocolumn,letterpaper]{article}

\usepackage{cvpr}      
\usepackage{array}
\usepackage{makecell} 
\usepackage{multirow}

\usepackage{graphicx}
\newcommand{\getimageinfo}[1]{%
  \begingroup
  \setbox0=\hbox{\includegraphics{#1}}%
  \typeout{^^JImage Info for #1:^^JWidth: \the\wd0^^JHeight: \the\ht0^^J}%
  \endgroup
}

\definecolor{cvprblue}{rgb}{0.21,0.49,0.74}
\usepackage[pagebackref,breaklinks,colorlinks,allcolors=cvprblue]{hyperref}

\def\paperID{12378} 
\def\confName{CVPR}
\def\confYear{2025}

\title{Emergence of Painting Ability via Recognition-Driven Evolution}

\author{
    Yi Lin$^{1,2}$, Lin Gu$^{3,4 \dagger}$, Ziteng Cui$^{4}$, Shenghan Su$^{1}$,  Yumo Hao$^{5}$, Yingtao Tian$^{6}$\\ Tatsuya Harada$^{3,4}$, Jianfei Yang$^{1 \dagger}$
    \\ $^{1}$Nanyang Technological University, $^{2}$National University of Singapore, 
   $^{3}$RIKEN,\\ $^{4}$The University of Tokyo,  $^{5}$Tianjin Academy of Fine Art, $^{6}$Sakana AI \\
}

\begin{document}

\getimageinfo{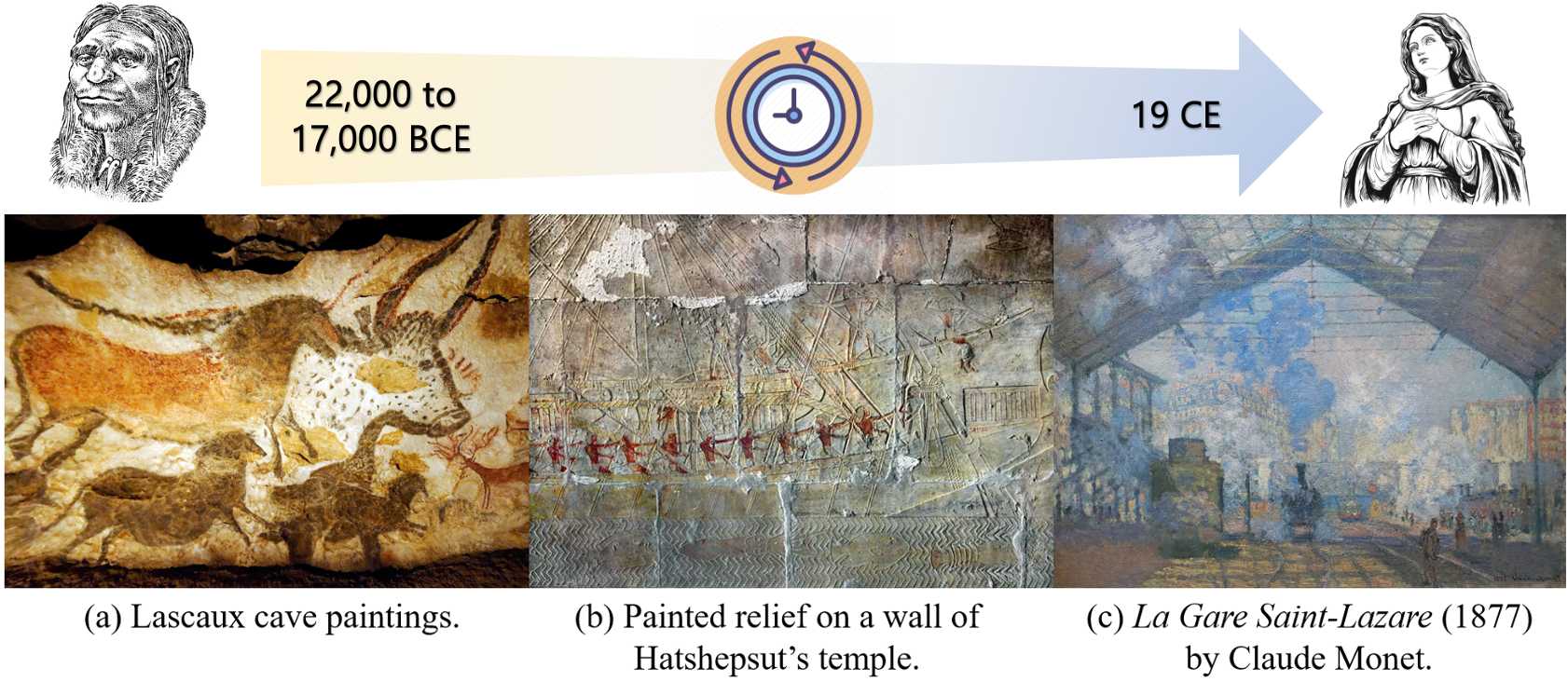}
\getimageinfo{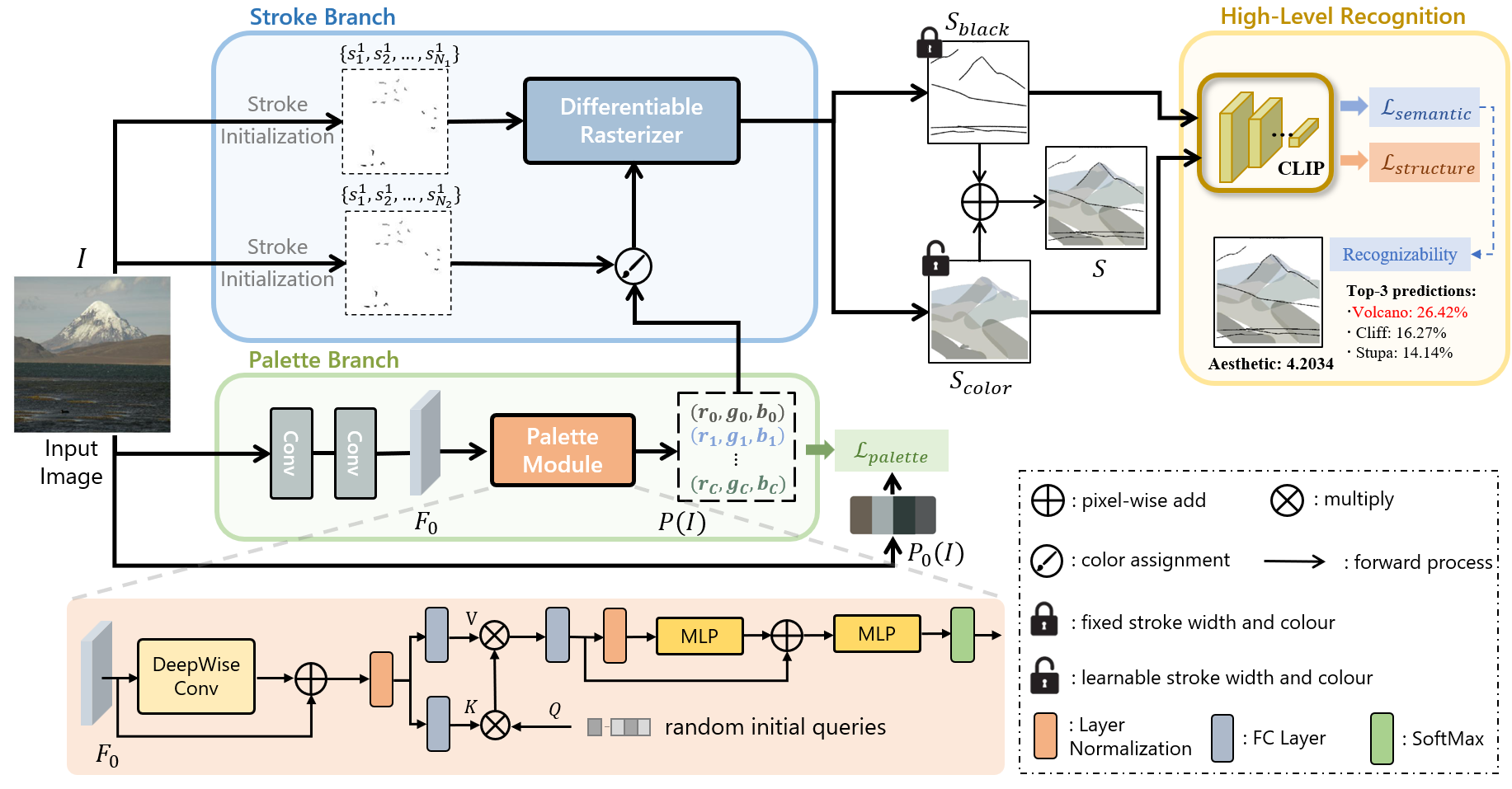}

\newcommand\mycommfont[1]{\footnotesize\rmfamily\textcolor{black}{#1}}
\maketitle
\renewcommand{\thefootnote}{}
\footnotetext{\hspace{-0.5cm}
$^\dagger$Co-corresponding authors. Lin Gu: \texttt{lin.gu@riken.jp}; Jianfei Yang: \texttt{jianfei.yang@ntu.edu.sg} \\
This work was conducted by the first author during her research internship at MARS Lab, NTU. }

\begin{abstract}
From Paleolithic cave paintings to Impressionism, human painting has evolved to depict increasingly complex and detailed scenes, conveying more nuanced messages. This paper attempts to emerge this artistic capability by simulating the evolutionary pressures that enhance visual communication efficiency. Specifically, we present a model with a stroke branch and a palette branch that together simulate human-like painting. The palette branch learns a limited colour palette, while the stroke branch parameterises each stroke using Bézier curves to render an image, subsequently evaluated by a high-level recognition module. We quantify the efficiency of visual communication by measuring the recognition accuracy achieved with machine vision. The model then optimises the control points and colour choices for each stroke to maximise recognition accuracy with minimal strokes and colours. Experimental results show that our model achieves superior performance in high-level recognition tasks, delivering artistic expression and aesthetic appeal, especially in abstract sketches. Additionally, our approach shows promise as an efficient bit-level image compression technique, outperforming traditional methods.
\end{abstract}

\section{Introduction}


 From the earliest days of human history, painting has served as a powerful means of communication.  Lascaux cave paintings, dating back over 17,000 years, predominantly depict animals such as cattle, wild horses, deer, and aurochs. Paleolithic humans used these illustrations to convey critical hunting knowledge, including prey types, hunting techniques, and migration patterns of certain animals. Ancient Egyptian art, exemplified by the reliefs in Queen Hatshepsut’s mortuary temple at Deir el-Bahari, depicts detailed trade scenes with Punt, including people, animals, and ships. According to al-Masudi, these depictions were so precise that enemies could be identified and targeted by destroying the corresponding reliefs. Unlike cave paintings, Egyptian art employed refined strokes and varied colours, leading to more accurate communication of information. Moving to modern times, Claude Monet's \textit{La Gare Saint-Lazare} captures the bustling energy of city life, portraying trains, steam, and crowds at a Paris station. His refined brushwork and expansive use of colours enhance both the liveliness and precision of the urban scene. These examples illustrate the evolution of human painting across three key dimensions: (1) \textbf{Scenario Complexity}, where scenes depicted in paintings become increasingly complex; (2) \textbf{Recognition Accuracy}, where conveyed information becomes more accurately; and (3)\textbf{ Abstraction and Colour Richness}, where the variety and detail as strokes and colours expand.

\begin{figure}
    \centering
    \includegraphics[width=\columnwidth]{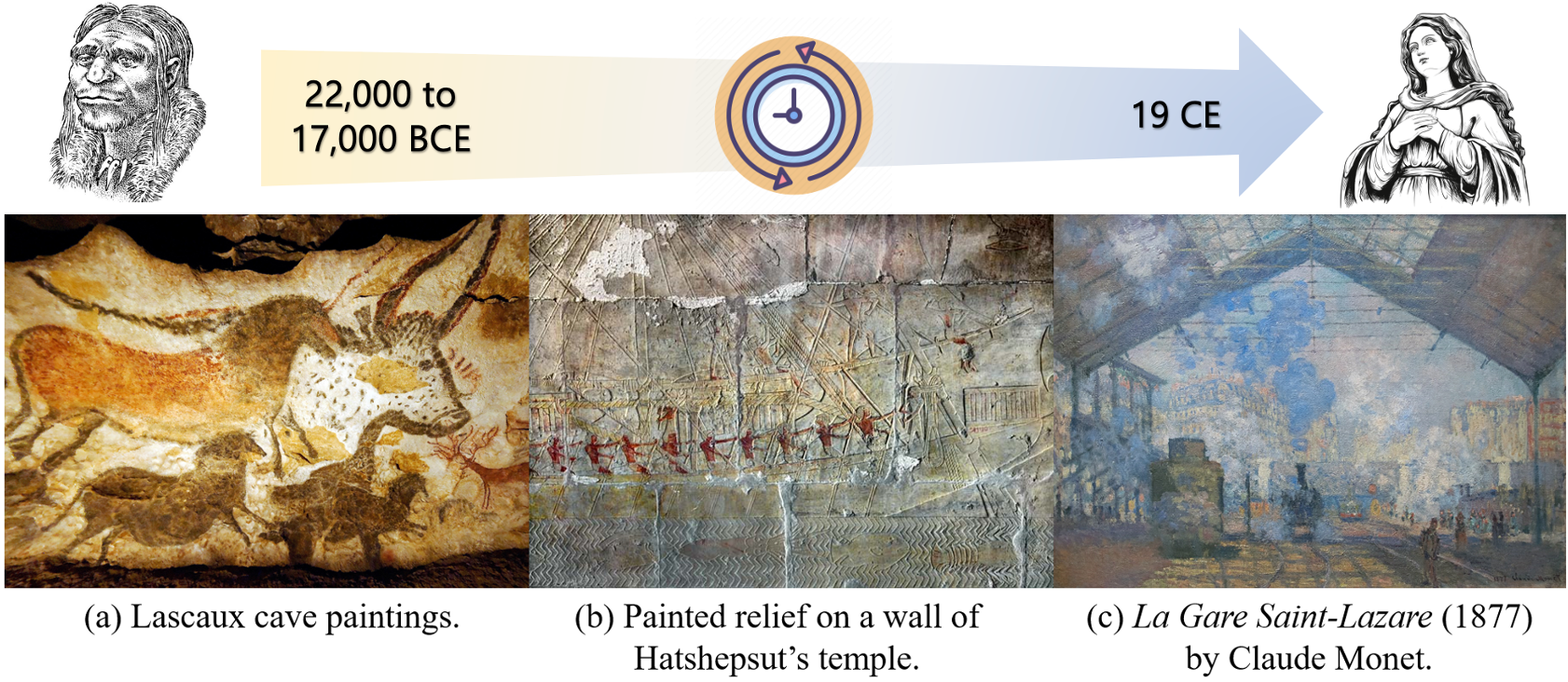}
    \vspace{-5mm}
    \caption{From cave paintings to Impressionism: the evolution of human painting is reflected in the increasing scenario complexity, enhanced recognition accuracy, and greater abstraction and colour richness. }
    \label{fig:human painting}
    \vspace{-3mm}
\end{figure}

\begin{figure*}
    \centering
    \includegraphics[width=1\linewidth]{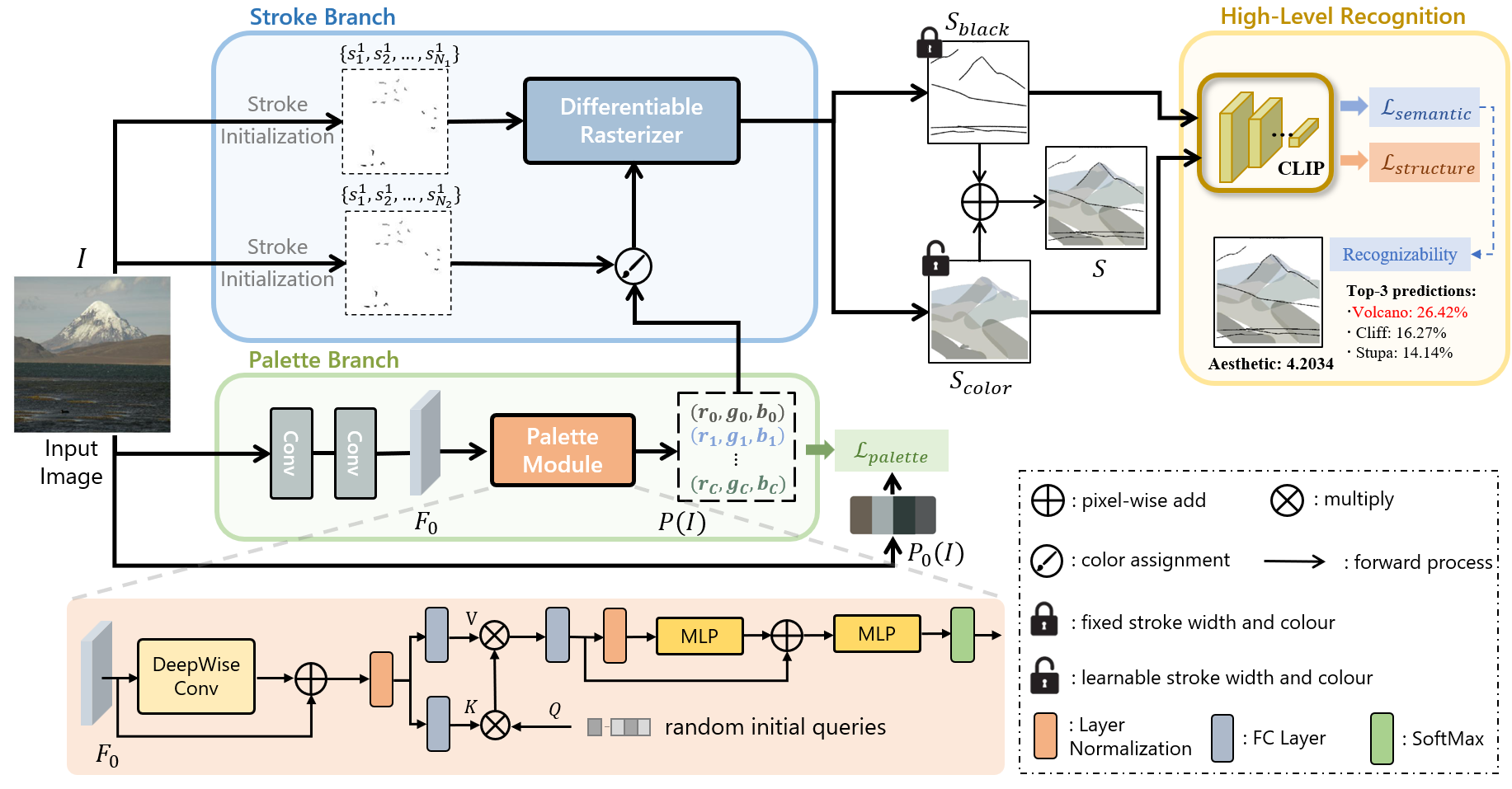}
    \vspace{-3mm}
    \caption{Overview of the painting model. The model generates human-like paintings using a Stroke Branch for stroke optimization and a Palette Branch for colour palette learning. A high-level recognition module is followed to evaluate the recognition accuracy of the generated painting. }
    \label{fig:model overview}
    \vspace{-3mm}
\end{figure*}

Evolution is ubiquitous, and human language provides a prominent example. Diachronic research ~\cite{zaslavsky2022evolution}~\cite{gibson2017color} demonstrates that languages continuously evolve to acquire new colour, leading to a more refined colour naming system. And it is hypothesised that language evolution is driven by the need for greater communication efficiency, allowing humans to convey precise information with fewer colour terms. Studies~\cite{chaabouni2021communicating}~\cite{su2023name} have shown that under evolutionary pressures, the structure of human colour naming emerges naturally, matching the basic colour terms theory \cite{berlin1991basic}. In this study, we investigate whether humans' ability to paint can emerge independently. As human painting evolves to achieve greater recognition accuracy, we define visual communication efficiency as the ability to convey information using minimal strokes and colours. Within this framework, our goal is to examine whether the ability to paint can autonomously develop under evolutionary pressures of visual communication efficiency.

Fig.~\ref{fig:model overview} illustrates the pipeline of our method, which transforms input images into human-like paintings represented by vector sketches~\cite{vinker2022clipasso}. Our proposed model consists of a painting model and a complexity estimation system. We use Bézier curves to represent each stroke in the vector sketch, with each stroke parametrized by its control points, RGB colour, and thickness. The painting model optimizes these parameters through two branches: the \textbf{Stroke Branch} and the \textbf{Palette Branch}.  The Stroke Branch, based on a Differentiable Rasterizer, iteratively updates the stroke parameters without relying on a specific sketch dataset for training. The Palette Branch uses a key-point localization strategy to learn a colour palette from the entire RGB space for rendering the strokes. At each iteration, the Stroke Branch generates a human-like painting by combining coloured strokes and black strokes with fixed width. The combined painting is passed to a high-level recognition module to evaluate the classification accuracy. We argue that evaluating recognition accuracy reflects visual communication efficiency under the constraints of restricted strokes and colours. Specifically, we minimize the semantic loss between the generated painting and the input image to achieve optimal classification accuracy.  For the strokes and colours constraints, we specify the number of strokes in the Stroke Branch and the number of colours in the Palette Branch. Through iterative optimization, our model achieves high recognition accuracy, showing how evolution occurs under the pressure of visual communication efficiency with more accurate information under the abstraction and colour richness constraints.

Besides the painting model, we introduce an estimation system to reflect the evolution of scenario complexity in painting, an aspect often overlooked in previous methods \cite{frans2022clipdraw}~\cite{vinker2022clipasso}~\cite{vinker2023clipascene}.  We propose two selection criteria: the \textbf{Category Criterion}, which identifies prominent object or scene categories common in historical artworks,  and the \textbf{Composition Criterion}, which selects images based on composition similarity to a referenced art collection.  We further propose a \textbf{scenario complexity estimator} that scores each image based on how easily an artist could capture the scene. Finally, we create a dataset tailored for painting generation and evaluate our model's performance using this dataset. 


Unlike methods that rely on explicit training \cite{li2019photo} \cite{song2018learning} \cite{kampelmuhler2020synthesizing} \cite{ramesh2022hierarchical} \cite{kim2022diffusionclip} \cite{zhang2023inversion} using human-painted images, our approach achieves the autonomous emergence of painting ability driven solely by evolutionary pressures. This reflects the machine's capability to independently understand and express its interpretation of the world.  Moreover, we were surprised to find that the paintings generated from our method share many similarities with human drawing styles. Professional artists have offered surprising evaluations, noting their complexity, technique, and scene selection, reminiscent of historical artistic styles such as Medieval church art. Additionally, our method demonstrates potential for low-bit-rate image compression, efficiently reducing image size while maintaining visual quality. It can also promote privacy and fairness by eliminating private characteristics like skin tone, texture, and other biases~\cite{yao2022improving}, helping prevent unfair discrimination in tasks such as image classification or detection.  

Our contributions can be summarized as follows:
\begin{itemize}
    \item  We propose a novel painting model where human-like painting ability autonomously emerges under evolutionary pressures, reflecting visual communication efficiency. 
    \item We introduce a system to estimate painting scenario complexity and create a dataset reflecting painting scenarios in different levels of complexity, facilitating the exploration of how painting scenes evolve over time. 
    \item Our approach delivers not only high artistic expression and aesthetic appeal, but also serves as an efficient low-bit-rate image compression technique, outperforming traditional methods in terms of visual quality and compression efficiency. 
\end{itemize}
\section{Related Work}

\subsection{Painting Generation}
Previous studies in image painting and sketch synthesis have approached painting generation as an image-to-image translation or style transfer problem~\cite{isola2017image}~\cite{zhu2017unpaired}~\cite{li2019im2pencil}~\cite{xu2021drb}. Most generative models use GAN-based architectures for sketch synthesis. \cite{song2018learning} proposed a shortcut cycle consistency loss to address the domain gap between vector sketches and reference images, using a supervised-unsupervised learning framework. \cite{kampelmuhler2020synthesizing} used sketch-image pairs as supervision to enhance realism and accuracy.  \cite{li2019photo} generates contour drawings capturing scene outlines, addressing limitations in boundary detection with a new dataset. \cite{chan2022learning} introduced an unpaired method for line drawing generation from photographs, using geometry and semantic losses to improve quality.  Other research~\cite{ramesh2022hierarchical}~\cite{kim2022diffusionclip}~\cite{zhang2023inversion} utilized diffusion models to generate high-quality images via iterative denoising. Although these methods produce impressive paintings in diverse styles, they depend on explicit painting or sketches datasets, limiting their ability to generate paintings independently, without relying on imitation of human painting styles. In contrast, our method enables the automatic emergence of painting ability, generating human-like sketches without relying on specific datasets.

Recent advances in differentiable rasterizers, such as DiffVG~\cite{li2020differentiable}, enables optimization-based vector sketch synthesis without relying on training datasets. Diffusion-based methods, such as VectoFusion~\cite{jain2023vectorfusion}, DiffSketch~\cite{wang2023diffsketching}, SVGDreamer~\cite{xing2024svgdreamer}, and VectorPainter~\cite{hu2024vectorpainter}, generate vector graphics from text prompts.  Other works, such as  CLIPDraw~\cite{frans2022clipdraw} , use pre-trained CLIP models to optimize stroke-based vector drawings by minimizing the difference between CLIP embeddings of generated sketches and input texts. For image-to-sketch generation, CLIPasso~\cite{vinker2022clipasso} propose a novel approch to synthesize object-level sketches that preserve both semantic and geometric features, while CLIPascene~\cite{vinker2023clipascene} focuses on scene-level sketching  at various levels of simplicity and fidelity. Recently, \cite{lee2023learning} proposed to learn geometry-aware sketch representations via CLIP-based perceptual loss. Specifically, they convert an image into colored Bézier curves that explicitly preserve the geometric properties of the images. Similarly, we use coloured Bézier curves to generate painting-like sketches that align with the input images. However, instead of using fixed stroke colours, we propose a Palette Branch that learns and iteratively updates a restricted colour palette by optimizing a colour-aware perceptual loss. Our model generates paintings that combine both colourized and black strokes, demonstrating strong recognition accuracy under evolutionary pressure. 

Although optimization-based painting generation methods eliminate the need for explicit training datasets, most still evaluate their models on datasets which overlook scenario selection in terms of scene complexity.  Existing sketch datasets, such as TU-Berlin~\cite{eitz2012hdhso}, Sketchy~\cite{sangkloy2016sketchy}, and ImageNet-Sketch~\cite{wang2019learning}, focus on everyday objects and have image-level categories annotation, while datasets like SketchyCOCO~\cite{gao2020sketchycoco} and SketchyScene~\cite{zou2018sketchyscene} expand to scene-level sketches but still fail to consider painting scenario selection. These datasets do not explore the diversity and complexity of historical painting scenarios, which is crucial for studying the evolution of hand-drawn sketches. Some datasets, like Im4Sketch~\cite{efthymiadis2022edge}, group objects by shape or semantic similarity but still focus on low- and high-level features rather than painting scenarios. Our proposed scenario complexity estimation system fills this gap via two criteria and one estimator, selecting images that align with historical painting scenes.


\subsection{Image Compression}
Image compression aims to represent an image with minimal bits while preserving visual quality and essential information. Generative models have shown great potential in this task, handling the balance between distortion and perception. \cite{agustsson2019generative} presents a GAN-based model that achieves high-quality images at low bit rates by using semantic labels to reduce storage.  \cite{yan2021perceptual} shows that high perceptual quality can be achieved with minimal MSE distortion.  \cite{agustsson2023multi} introduces a generative method with explicit control over the distortion-realism trade-off, allowing for varying reconstruction detail.  Recent work, such as EGIC~\cite{korber2025egic}, also effectively traverses the distortion-perception curve.

While compression minimizes file size, sketch synthesis creates minimalistic representations that retain key features. Both tasks prioritize efficiency—whether in bits (compression) or strokes (sketch synthesis). Building on Marr’s insight ~\cite{man1982computational}, the primal sketch model~\cite{guo2007primal} shows that integrating structural and textural components can provide a concise representation, achieving high compression (over 1:25) while preserving texture perception. Similarly, our method compresses images by vector representations with high abstraction levels and extremely restricted colour richness, reducing complexity while maintaining recognition accuracy.

\section{Painting Model}

\subsection{Overall Architecture}
\label{section3.1structure}
An overview of our proposed painting model is presented in Fig.~\ref{fig:model overview}. Given an input image \( I \in \mathbb{R}^{H \times W \times 3} \), our goal is to synthesize a corresponding painting \( S \) that preserves recognition accuracy. We represent painting as a vector sketch using a set of Bézier curves, each parameterized by a set of four control points  \( \{ p_j \}_{j=1}^4 \). A painting \( S \) is composed of \( N \) strokes, denoted as \( (s_1, s_2, \ldots s_N) \). For each stroke \( s_i \), the parameters to optimize positions of four control points \( \{ (x_{j}, y_{j}) \}_{j=1}^4 \), the RGB colour \( (r_i, g_i, b_i) \), and the stroke thickness \( w_i \). 

Our painting model comprises two main components: a Stroke Branch and a Palette Branch. The Stroke Branch is primarily built upon diffVG~\cite{li2020differentiable}, a differentiable rasterizer that allows for the backpropagation of gradients through the rasterization process. Takes the original image \( I \) as input, after stroke intialization, the Stroke Branch generates a painting \( S \) by rendering one black-and-white sketch  \( S_{\text{black}}\) and one coloured sketch \( S_{\text{colour}}\) through the differentiable rasterizer and integrating both. During the rendering process, the colour and width are fixed for  \( S_{\text{black}}\), while the colour and width of  \( S_{\text{colour}}\) are optimizable. The Palette Branch takes the original image \( I \) and Reference Palette Queries as inputs and generates a colour palette \( P(I) \in \mathbb{R}^{C \times 3} \) by localizing the 3D position within the entire RGB colour space. At each iteration, colours learned in \( P(I) \) are assigned to the strokes. The generation of  \( P(I) \) is optimized by minimizing the colour-aware perceptual loss between the output and a Reference colour Palette  \( P_0(x) \in \mathbb{R}^{C \times 3} \), which is a set of RGB values obtained during stroke initialization based on the attention map.

Finally, paintings generated by our painting model is passed to a high-level recognition module to compute the classification accuracy.

\begin{figure}
    \centering
    \includegraphics[width=1\linewidth]{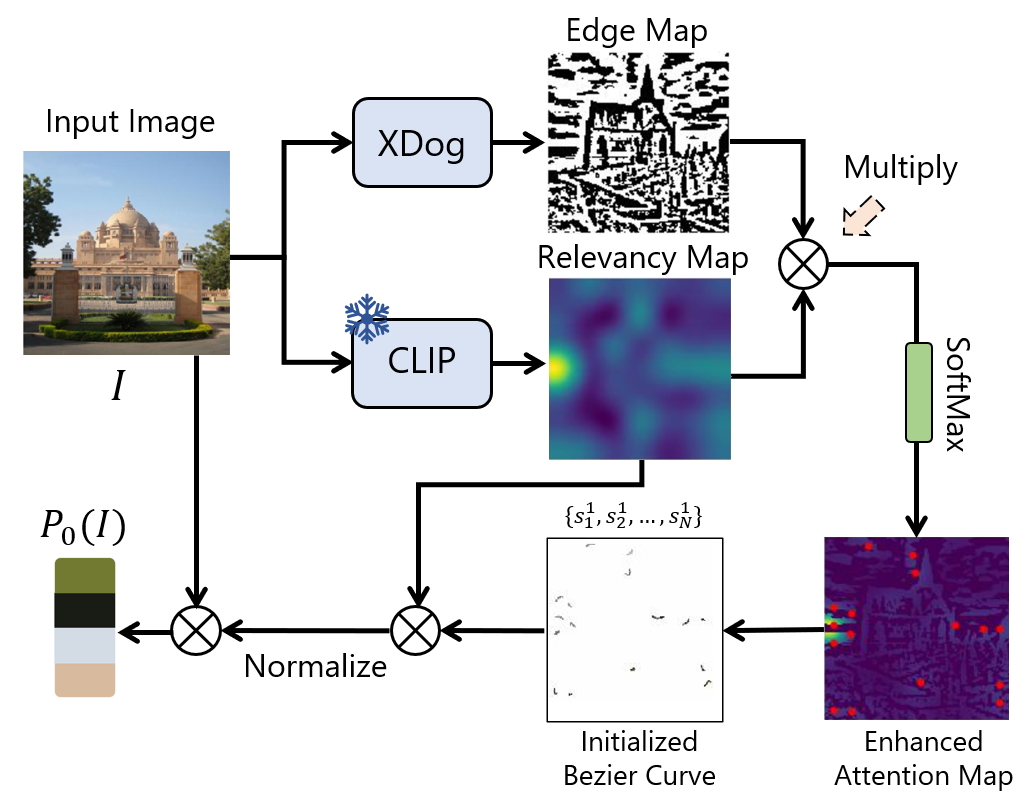}
    \caption{The process of stroke initialization and the generation of the Reference Colour Palette.}
    \label{fig:model detail}
     \vspace{-5mm}
\end{figure}

\subsection{Stroke Branch}
\label{section3.2stroke}
\subsubsection*{Stroke Initialization} We follow the approach in CLIPasso~\cite{vinker2022clipasso}, using the pre-trained Vision Transformer model ViT-B/32 for global context modeling~\cite{alexey2020image} of the input image. As shown in Fig.~\ref{fig:model detail}, a saliency map is extracted from the self-attention heads and multiplied by the edge map generated using XDoG ~\cite{winnemoller2012xdog}, producing an enhanced attention map. Subsequently, a softmax function is applied to normalize the enhanced attention map, assigning higher probabilities to pixels in significant regions. This map is then normalized with a softmax function, assigning higher probabilities to important regions. The highest-weight point in the normalized map is selected as the first control point for the initial stroke, and three additional control points are randomly sampled within a small neighborhood around it.

During stroke initialization, we design a Reference Colour Palette \( P_0(I) \) for colour optimization in the Palette Branch. To obtain \( P_0(I) \), we convert the initialized stroke to a raster image, multiply it by the relevancy map to highlight the overlapping areas between strokes and attention regions, and normalize the result to ensure the pixel values sum to one. We then multiply this with the input image to compute the RGB values at the corresponding stroke locations. This process is repeated for all strokes, generating a Reference Colour Palette with \( C \) distinct RGB values. By focusing on the most salient regions, \( P_0(I) \) captures the most representative colours in the image.

\subsubsection*{Stroke Optimization} After stroke initialization, we have two sets of stroke parameters: \( S_{\text{black}} = \{ s_1^1, s_2^1, \ldots, s_{N_1}^1 \} \) and \( S_{\text{colour}} = \{ s_1^2, s_2^2, \ldots, s_{N_2}^2 \} \). Unlike diffVG and other diffVG-based sketch synthesis methods~\cite{wang2023diffsketching}~~\cite{frans2022clipdraw}~\cite{vinker2022clipasso}~\cite{vinker2023clipascene}, which render a single sketch at each iteration, our Stroke Branch optimizes two sets of strokes and produce a black sketch and a coloured sketch, simultaneously.  For \( S_{\text{black}} \), only the control point positions are optimized, while the stroke colour remains black and the stroke width is fixed at 1.0. For \( S_{\text{colour}} \), we optimize both stroke colour and width to mimic the colour blending effect in human paintings, such as watercolour and ink. The opacity of the strokes is fixed at 0.5, allowing the blending of multiple coloured strokes to simulate light or heavy brushstroke intensity within the painting.

To reflect the visual communication efficiency in our painting model, we compute the semantic loss \( L_{\text{semantic}} \)  to minimize the distance between the high-level semantic embedding of the generated painting and the input image. To provide consistency between the low-level features of the generated painting and that of the input image, we use a structural loss \( L_{\text{structure}} \) to enforce geometric similarity in the painting. Please refer to  Sec.~\ref{section3.4recognition} for more details. 

\subsection{Palette Branch}
\label{section3.3palette}For the colouring of strokes, the Palette Branch learns representative colours through the localization of 3D spatial coordinates within the entire RGB colour space, which is a strategy utilized in various key-point detection tasks~\cite{carion2020end}~\cite{li2021pose}~\cite{cui2022you}. Different from these work, our Palette Module processes the input image \( I \) alongside a learnable Reference Palette Queries \( Q \in \mathbb{R}^{C \times d} \) to produce an output colour palette \( P(I) \in \mathbb{R}^{C \times 3} \), where \( C \) denotes the number of colours. The queries \( Q \) is a set of \( C \) learnable embeddings, each of dimension \( d \), corresponding to the RGB values of the automatically learned colours.

The Palette Branch consists of positional encoding, a colour Transformer, and a colour coordinate projector. Given an RGB input image \( I \), we first apply two convolutional layers to produce a \( d \)-dimensional feature map \( F_0 \in \mathbb{R}^{\frac{H}{4} \times \frac{W}{4} \times d} \). Using depth-wise convolution, we apply positional encoding to  \( F_0 \), resulting in position-encoded features  \( F_0' \).  \( F_0' \) is flattened and passed to the transformer blocks. During cross-attention computation, the keys \( K \) and values \( V \) are derived from \( F_0 \) through two fully connected (FC) layers. The outputs of the cross-attention block are processed by a multi-layer perceptron (MLP) with two FC layers, a GELU ~\cite{hendrycks2016gaussian} activation, and a residual connection, yielding an embedding based on the learned query \( Q \). 

Finally, the output embedding is projected into the final colour palette across the entire RGB colour space via the colour coordinate projector, another MLP with two fully connected layers and GELU activation . After applying a sigmoid activation, we obtain the final colour palette  \( P(I) \), comprising \( C \) RGB values assigned to the strokes in the rendering process.

To ensure \( P(I) \)  represent the RGB colour space with a constrained number of colours,  we introduce a colour-aware perceptual loss \( L_{\text{colour}} \) for effective colour optimization. \( L_{\text{colour}} \) computes the channel-wise mean squared error between the Reference colour Palette \( P_0(I) \) and the predicted colour palette \( P(I) \) predicted.

\subsection{Recognition Module}
\label{section3.4recognition}
After the painting model produces the colourized painting, we evaluate its visual communication efficiency through a high-level recognition task. Here, we introduce a Recognition Module and compute the semantic loss. 

The semantic loss \( L_{\text{semantic}} \) is central to our method, as it reflects the recognition accuracy of the generated painting.  It is computed by measuring the cosine similarity between the high-level embeddings of the generated painting and the input image. Minimizing this loss ensures the painting accurately captures the semantic content of the input, aiding in more accurate recognition. 

Using the output embedding from a pre-trained CLIP image encoder, which captures global semantic information,   \( L_{\text{semantic}} \) is defined as:
\[
L_{\text{semantic}} = \sum_l \text{cosine}(\phi(I), \phi(S))
\]

Additionally, we introduce a structural loss   \( L_{\text{structure}} \) to preserve low-level features, such as shapes and contours. While it doesn't directly affect recognition, it plays a supplementary role by maintaining structural integrity and geometric consistency.   Based on the same pre-trained CLIP encoder,  \( L_{\text{structure}} \) measures the L2 distance between the low-level embeddings of the generated painting and the input image: 
\[
L_{\text{structure}} = \sum_l L_2(\phi_l(I), \phi_l(S))
\]
where \( \phi \) is the CLIP image encoder and \( \phi_l \) is the CLIP image embedding at layer \( l \).

The final loss objective is defined as:
\[
L = \lambda_1 L_{\text{structure}} + \lambda_2 L_{\text{semantic}} + \lambda_3 L_{\text{colour}}
\]
where \( \lambda_1, \lambda_2, \) and \( \lambda_3 \) are set to 1.0, 1.0, and 1.0 as hyperparameters.

\section{Scenario Complexity Estimation}

\begin{figure}
    \centering
    \includegraphics[width=0.75\linewidth]{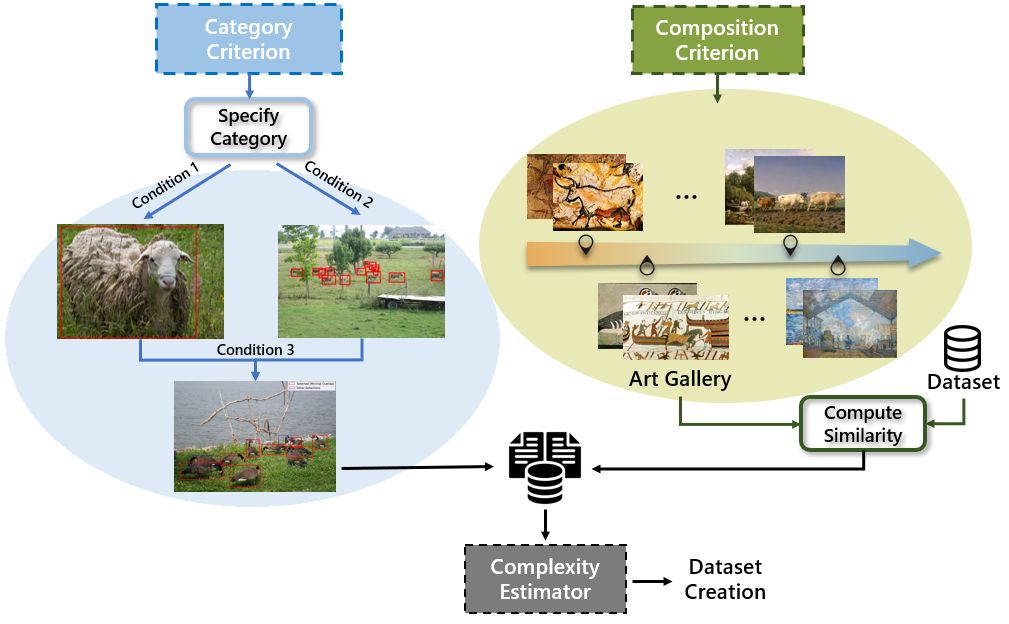}
    \vspace{-3mm}
    \caption{Overview of the scenario complexity estimation system.}
    \label{fig:estimator}
    \vspace{-5mm}
\end{figure}

Existing sketch datasets often neglect scenarios commonly depicted in artistic paintings. To address this gap, we propose a Scenario Complexity Estimation System. Images with higher scenario complexity typically involve more objects, finer details, and intricate spatial relationships, making them harder to depict accurately. Building on art painting knowledge, we incorporates two criteria and one estimator. The Category Criterion filters images based on specific categories, while the Composition Criterion selects images with high-level features similar to those in representative paintings. Inspired by ~\cite{sangkloy2016sketchy}, we trained a scenario complexity estimator to estimate scene complexity. This system allows us to select images from natural datasets that match artistic scenarios of varying complexity, helping create a dataset to explore the emergence of machine painting ability. An overview of this proposed system is shown in Fig.~\ref{fig:estimator}.

\subsection{Category Criterion}
\label{sec4.1category}
We utilise three widely adopted datasets in image classification and object detection to demonstrate our system and validate the performance of our model: ImageNet~\cite{deng2009imagenet}, COCO~\cite{lin2014microsoft}, and PASCAL VOC Dataset~\cite{everingham2010pascal}.

With the assistance of a professional artist, we collected an art gallery containing 60 paintings from Neolithic cave paintings to modern art and identified category lists for each dataset. While these lists guide image selection, an image may contain multiple instances or categories, leading to ambiguity for high-level recognition task. To assess the scenario complexity based on the objects and details in an image, we define three conditions: 
\begin{itemize}
    \item \textbf{Condition 1} (Small and Numerous Bounding Boxes): Images with at least ten bounding boxes labeled with a specific category and areas under 5\% of the image area, ensuring dense scenarios with many small objects.
    \item \textbf{Condition 2} (Large Bounding Boxes): At least one bounding box of a specific category has an aspect ratio greater than 50\% of the image area, ensuring the more dominant elements contribute to scene complexity.
    \item \textbf{Condition 3} (Minimal Overlapping): The Intersection over Union (IoU) between any two bounding boxes in the image must be less than 0.1, ensuring the scenario is clearly defined with least ambigutity.

\end{itemize}

\subsection{Composition Criterion}
\label{sec4.2comparison}
The Composition Criterion serves as a complement to the Category Criterion, evaluating the alignment of images with the visual composition of human paintings. Specifically, we use the art gallery collected in Sec.~\ref{sec4.1category} as a reference template. Our approach utilizes the image encoder of a pre-trained CLIP model to assess the similarity between images in the dataset and these reference paintings, with cosine similarity used as the metric for comparison. For each painting in the gallery, we select the top-10 images with the highest similarity, ensuring that only images that closely resemble the artistic compositions are included. 

\subsection{Scenarios Complexity Estimation}
\label{sec4.3sketchability}
Trained on the Sketchy dataset~\cite{sangkloy2016sketchy}, the estimator classifies images into five complexity levels: \{0, 1, 2, 3, 4\}, where 0 is "simplest" and 4 is "intricate." The complexity estimator functions as a classifier, consisting of an image encoder derived from the pre-trained CLIP model and a classification head. Based on the predicted score,  we filter images and create a dataset of around 500 images across over 70 categories, including 260 images from ImageNet and 250 images from the COCO and the PASCAL VOC dataset. Refer to the supplementary materials for more details and statistics about this dataset.

\section{Experiments}


\subsection{Quantitative Evaluation}
In this section, we evaluate the performance of our proposed model using two metrics: recognition accuracy and aesthetic score. Our experiments demonstrate that our model excels in the high-level recognition task while generating visually appealing and artistically expressive human-like paintings. 

We assess recognition accuracy as a measure of the visual communication efficiency of the generated painting, focusing on how effectively the model conveys information with minimal strokes and colours. Specifically, recognition accuracy indicates how well a painting can be identified by a machine as belonging to the same category as the original image. To achieve this, we perform zero-shot classification using a pre-trained CLIP ViT-32/B~\cite{radford2021learning}, calculating the cosine similarity between the image embedding of the generated sketch and the text embedding of text prompts defined by "\textit{A sketch of a(n) {classname}}". 

While recognition accuracy predicts the category of the painting, the aesthetic score assesses its visual appeal. Aesthetic score is a reflection of ratings given by individuals when evaluating the visual appeal of the painting. We compute this score using an aesthetic indicator ~\cite{github_project}. This indicator is trained on the LAION ~\cite{schuhmann2022laion} dataset and features a CLIP ViT-14/L backbone and a multi-layer perceptron (MLP).

\begin{table}[t]
\caption{Top-1 and Top-3 average recognition accuracy on the generating paintings on 190 images across 40 categories from the ImageNet dataset. Best results shown in bold.}
\label{tab:accuracy}
\renewcommand\arraystretch{2.0}
\resizebox{\linewidth}{!}{
\begin{tabular}{c|c|c c c c} \Xhline{1.1pt}

\multirow{2}{*}{\textbf{\begin{tabular}[c]{@{}c@{}}Avg. Recognition \\ Accuracy\end{tabular}}} & \multirow{2}{*}{\textbf{Metric}}
& \multicolumn{4}{c}{\textbf{Number of Strokes}} \\ \cline{3-6}  
& & 4 & 8 & 16 & 32 \\ \hline     
\( S_{\_} \) &{\makecell[c]{Top-1 \\
Top-3}}& {\makecell[c]{1.58\% \\
7.89\%}}& {\makecell[c]{5.26\% \\
16.32\%}}& {\makecell[c]{11.58\% \\
30.00\%}}& {\makecell[c]{29.48\% \\
50.53\%}}\\ \hline  
\( S_{\text{colour}} \) &{\makecell[c]{Top-1 \\
Top-3}}& {\makecell[c]{3.16\% \\
7.89\%}}& {\makecell[c]{2.63\% \\
10.00\%}}& {\makecell[c]{7.37\% \\
20.53\%}}& {\makecell[c]{20.53\% \\
39.47\%}}\\ \hline  
\( S_{\text{black}} \)&{\makecell[c]{Top-1 \\
Top-3}}& {\makecell[c]{2.63\% \\
6.32\%}}& {\makecell[c]{2.11\% \\
10.53\%}}& {\makecell[c]{9.47\% \\
23.68\%}}& {\makecell[c]{24.74\% \\
48.42\%}}\\ \hline
 CLIPasso&{\makecell[c]{Top-1 \\
Top-3}}& {\makecell[c]{2.63\% \\
7.37\%}}& {\makecell[c]{3.16\% \\
12.63\%}}& {\makecell[c]{10.53\% \\
24.21\%}}&{\makecell[c]{27.89\% \\
48.42\%}}\\ \Xhline{1.1pt}
\end{tabular}}
\vspace{-3mm}
\end{table}

\subsubsection*{Evaluate Recognition Accuracy}
Guiding by the scenario complexity estimation system, we selected 190 images across 40 categories from ImageNet to generate human-like paintings and evaulate their recognition accuracy. Table~\ref{tab:accuracy} shows the top-1 and top-3 recognition accuracy of \( S_{\text{black}} \), \( S_{\text{colour}} \), and the painting \( S \), under four levels of abstraction (4, 8, 16, and 32 strokes). The palette is restricted to 16 colours, but for 4- and 8-stroke cases, only 4 and 8 colours are learned, respectively. As shown in the table, as stroke number increases, the recognition accuracy of both \( S_{\text{black}} \) and \( S_{\text{colour}} \) increases, indicating that more strokes result in more accurate information conveyance.

We also compare our method with CLIPasso~\cite{vinker2022clipasso}. For fair comparison, we use the same number of strokes, stroke width, and number of iterations in CLIPasso as in \( S_{\text{black}} \), without applying foreground extraction masks. As shown in the last row in Table~\ref{tab:accuracy}, the recognition accuracy of CLIPasso is close to that of \( S_{\text{black}} \) from our method, which is likely because optimization of \( S_{\text{black}} \) is similar to that of CLIPasso, integrating both semantic and structural losses. However, by incorporating optimized coloured strokes, our method achieves higher accuracy in the final painting \( S_{\text{colour}} \), particularly top-3 accuracy of 50.53\% with 32 strokes, compared to 48.42\% of CLIPasso. These experiment results reflect the emergent human-like painting ability of our model, emphasizing how painting evolution is driven by visual communication efficiency and leads to more accurate visual representations with limited strokes and colours.

\subsubsection*{Evaluate Aesthetic Score}
\begin{figure}
    \centering
    \includegraphics[width=1\linewidth]{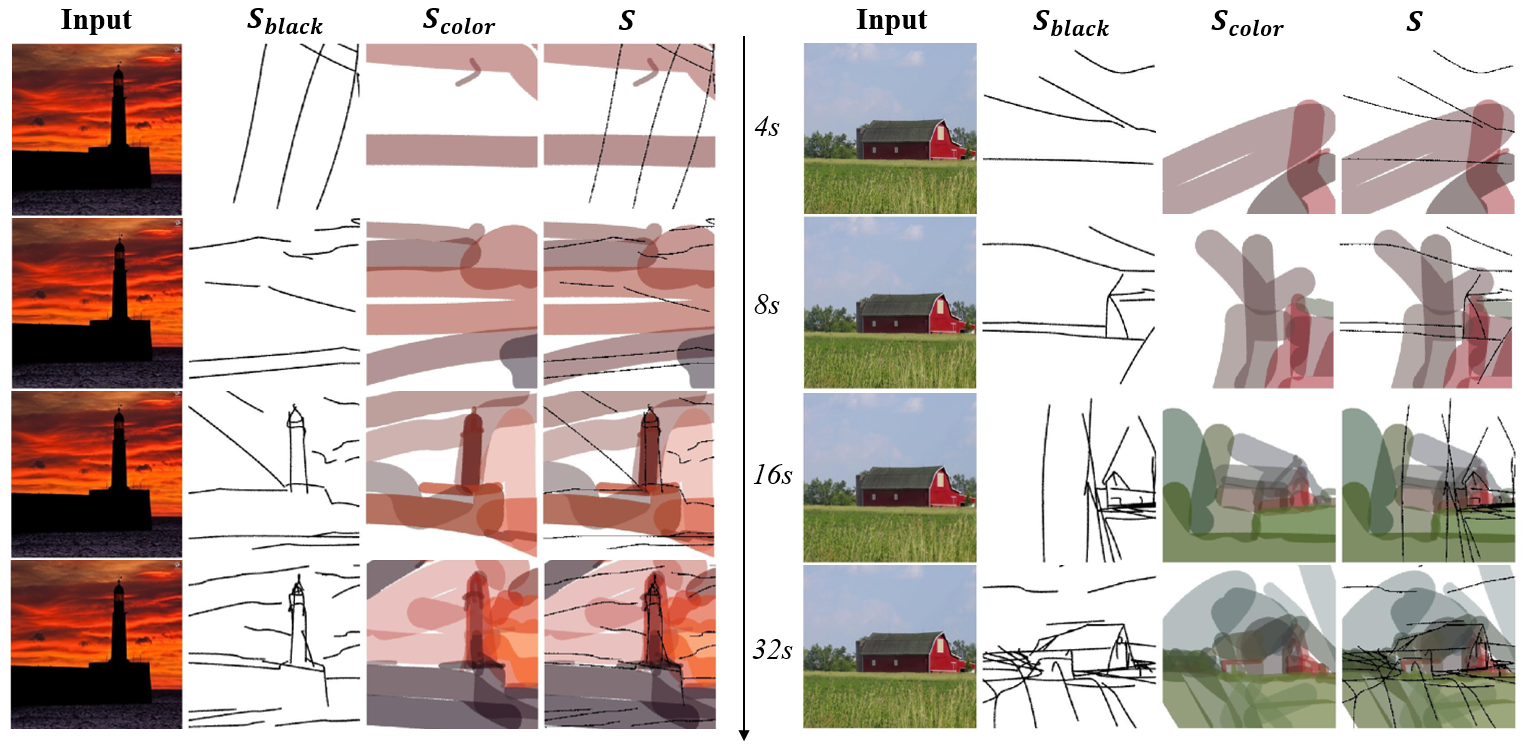}
    \vspace{-5mm}
    \caption{Generated paintings with different numbers of strokes. From top to bottom, both \( S_{\text{black}} \) and \( S_{\text{colour}} \) is produced using 4, 8, 16, and 32 strokes.}
    \label{fig:num_stroke}
\end{figure}

As shown in Table~\ref{tab:score}, we evaluate the aesthetic score on 100 images across 35 categories selected from the COCO and PASCAL VOC datasets, varying the number of strokes. The final paintings from our method, combining both black and coloured strokes, achieve the highest aesthetic scores, outperforming both black and coloured sketches. This demonstrates that our method can produce human-like paintings that express aesthetic appeal. Also, we find that as the number of strokes increases, the aesthetic score improves. Figure~\ref{fig:num_stroke} provides a straightforward visualization. When the number of strokes is increased from 4 to 32, both \( S_{\text{black}} \) and \( S_{\text{colour}} \) change from abstract and simple representations to more detailed and recognizable paintings, which is consistent with the results in Table~\ref{tab:accuracy}. Besides, we notice that the coloured strokes not only enhance structural features but also add texture details to the black-and-white sketch. For example, from the painting in the left part of Figure~\ref{fig:vis}, the giraffe's spots and the dark feathers of the bird's wings and eyes are highlighted by the coloured strokes, while the black strokes fail to capture these textural features. Moreover, we can observe a higher recognition accuracy in paintings which combine both the black and coloured strokes. These results illustrate that our model can automatically emerge human-like painting ability, where adding more strokes detail and colour richness results in more aesthetically pleasing paintings. 



\begin{table}[t]
\caption{Average aesthetic score on 100 images across 35 categories from the CoCo and PASCAL VOC dataset.}
\label{tab:score}

\renewcommand\arraystretch{1.1}
\resizebox{\linewidth}{!}{%

\begin{tabular}{c|c|c|c|c}

\Xhline{1.1pt}
          \small{\textbf{Avg. Aesthetic Score}}& \multicolumn{4}{c}{\small{\textbf{Number of Strokes}}} \\ \cline{2-5}
        & 4& 8& 16& 32\\ \hline
        \( S\)& 3.932& \multicolumn{1}{c|}{4.179} & 4.301& 4.299\\ \hline
        \( S_{\text{black}} \)& 3.849& 3.921& 4.179& 4.103\\ \hline
        \( S_{\text{black}} \)& 3.673
& 3.845
& 3.698& 4.004\\ \hline
 \( I\)& \multicolumn{4}{c}{4.628}\\ 
\Xhline{1.1pt}
\end{tabular}}
\vspace{-3mm}
\end{table}

\begin{figure*}
    \centering
    \includegraphics[width=1\linewidth]{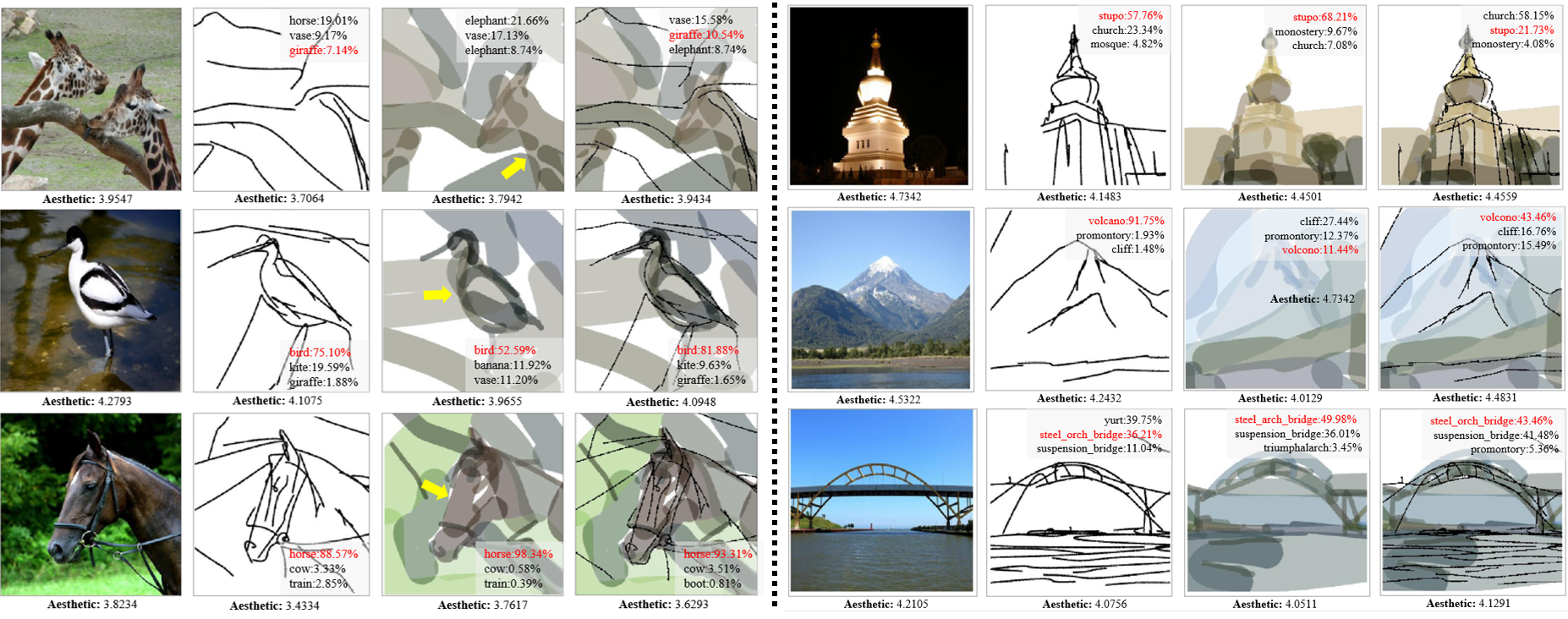}
    \vspace{-4mm}
    \caption{Top-3 predictions and the corresponding recognition accuracy of the generated paintings. Right: examples which demonstrate that coloured strokes not only enhance structural features but also add more texture details to the black sketch. Left:  examples which reveal that painting may prioritize simplicity for better visual communication efficiency when low-level features such as contour are dominant in the images.}
    \label{fig:special_case}
\end{figure*}


Interestingly, we found that while colour enhances aesthetic appeal, it doesn't always improve recognition accuracy. As shown in Figure~\ref{fig:special_case}, in some cases, \( S_{\text{black}} \) yielded higher recognition accuracy than the final paintings \( S\), despite the latter having higher aesthetic scores. This suggests that visual communication efficiency may prioritize lower abstraction levels and less colour richness, emphasizing structural features like shapes and contours for better recognition. 



\begin{figure}
    \centering
    \includegraphics[width=1\linewidth]{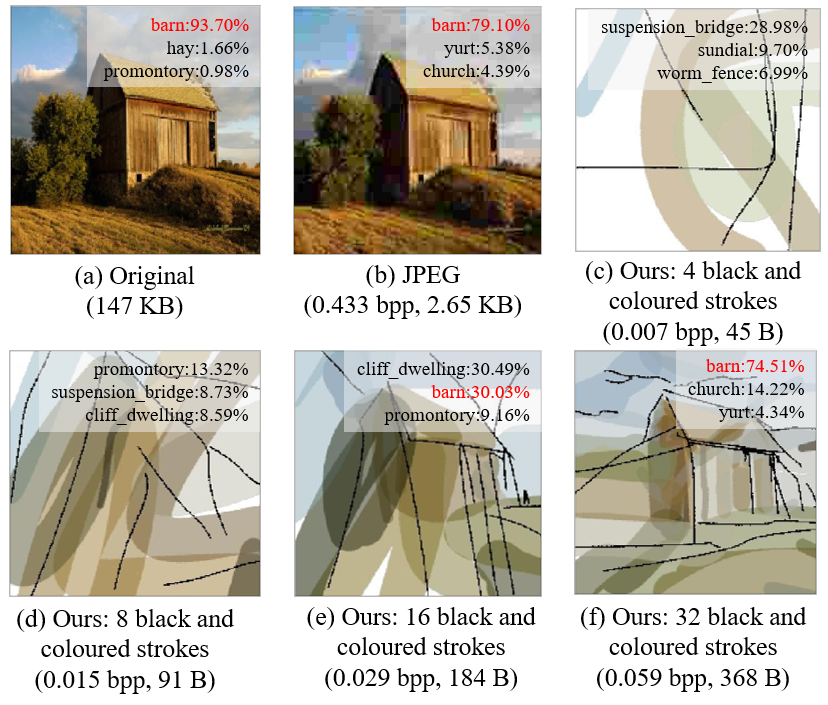}
    \vspace{-4mm}
    \caption{Comparison of image quality and recognition accuracy among the original image, JPEG image, and paintings generated by our model. (c) - (d) show the results for varying stroke counts from 4 to 32, for both black and coloured strokes. }
    \label{fig:compression}
    \vspace{-4mm}
\end{figure}

\subsection{Image Compression Performance}

Unlike traditional raster-based compression methods, our approach compresses the image at a much lower bit rate while maintaining visual recognition performance. In Figure~\ref{fig:compression}, we demonstrate the potential of our painting model in low-bit-rate image compression, outperforming traditional methods. Specifically, as shown in the comparison among the original image (a), JPEG (b), and the sketches generated with varying levels of abstraction (c-f), our method achieves significantly lower bits while maintaining competitive recognition accuracy. For instance, using 32 black strokes and 32 coloured strokes (Figure~\ref{fig:compression} (f)), our method reaches only 0.059 bits per pixel (bpp), which is comparable to JPEG compression at 0.433 bpp  (Figure~\ref{fig:compression} (b)). Despite this drastic reduction in data size, our approach achieved an accuracy of 74.51\% which is comparable to 79.10\% achieved in JPEG. The combination of black and coloured strokes effectively represents the key features of the image, making our approach a highly efficient solution for vector-based image compression. More comparison results with other traditional compression methods~\cite{floyd1976adaptive}~~\cite{gervautz1988simple}~\cite{heckbert1982color} are shown in the discussion section.  

\section{Discussion}

\begin{figure}
    \centering
    \includegraphics[width=1\linewidth]{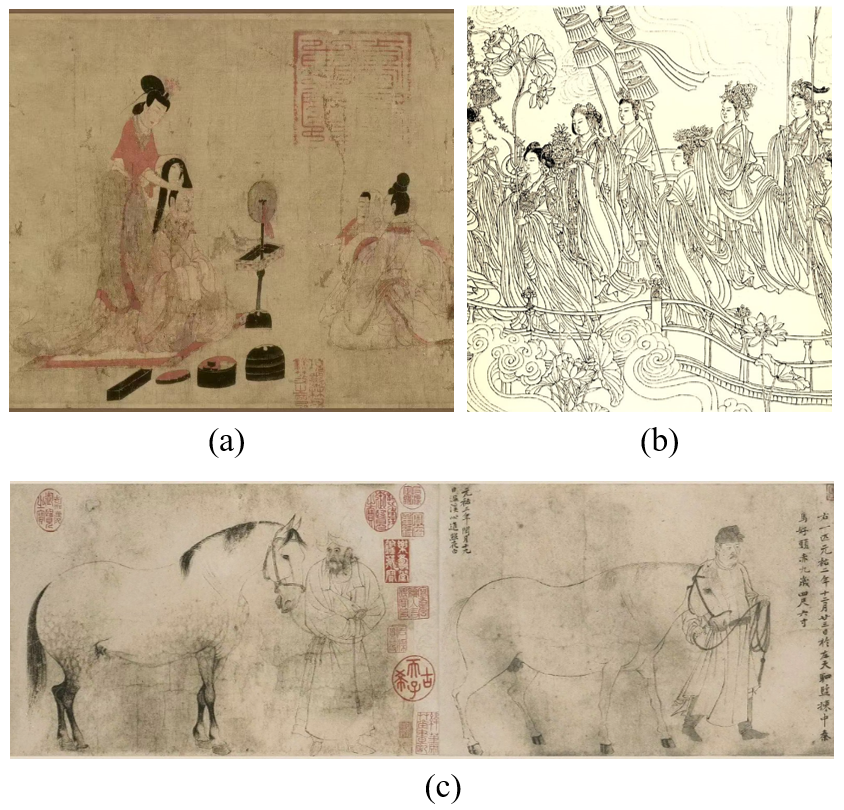}
    \vspace{-4mm}
    \caption{Line drawing in ancient China. There is a slight transition of the painting technique from flat outlines to a more stereoscopic depiction of the contour. (a) \textit{Admonitions of the Court Instructress} by  Gu Kaizhi (c.345-c.406 CE) in the Eastern Jin Dynasty. (b) Part of the \textit{Eighty-Seven Immortals} scroll by Wu Daozi (c.680-c.759 CE) in the Tang Dynasty. (c) Part of the \textit{Five Horses} by Li Gonglin (c.1049-c.1106 CE) in the Song Dynasty.}
    \label{fig:linedrawing}

\end{figure}

\begin{figure}
    \centering
    \includegraphics[width=1\linewidth]{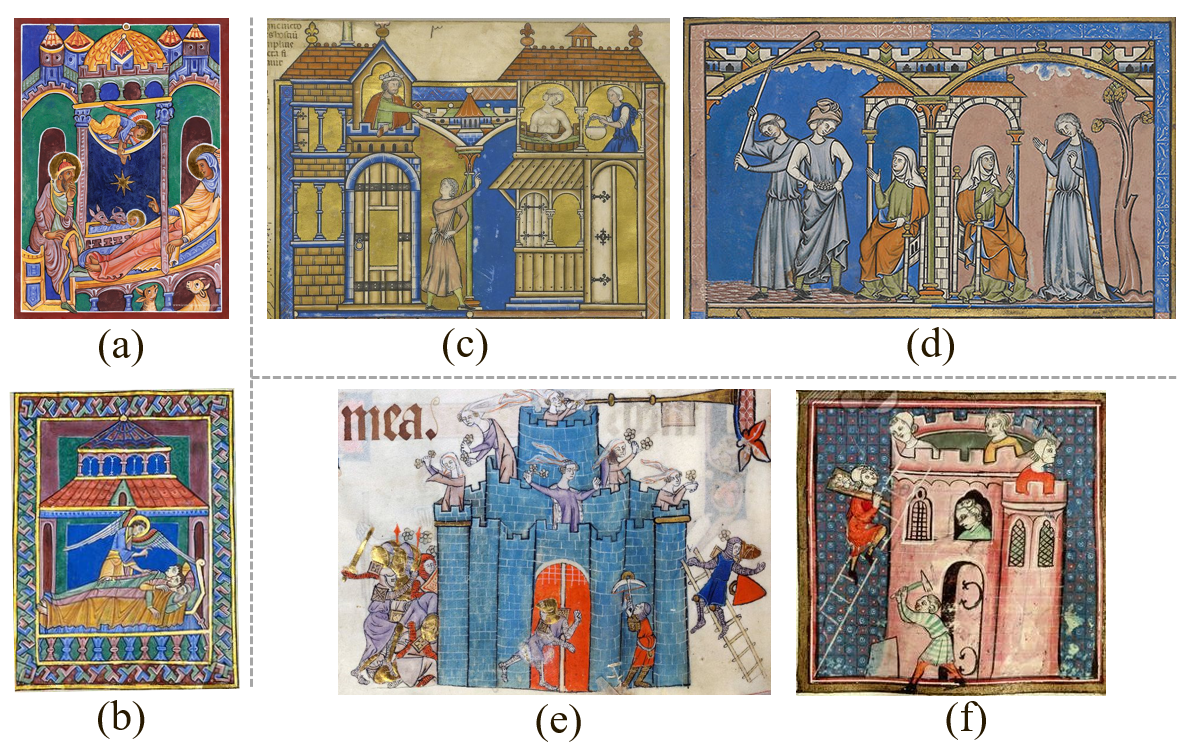}
    \vspace{-4mm}
    \caption{Illustration of Medieval Painting Art.(a)-(b) Illustrations to the \textit{St.Albans Psalter} (12th   Century). (c)-(d) Illustration to the \textit{Crusader Bible} (1240s). (e) Illustration to the \textit{Luttrell Psalter} (1320-1345). (f) Illustration to the Bible Historiale of Guyart des Moulins (14th Century).}
    \label{fig:medieval}
    \vspace{-3mm}
\end{figure}
The most remarkable aspect of our approach lies in the model's ability to emerge with painting ability independently, without directly mimicking human paintings. This emergent painting ability is particularly intriguing when contextualized within the framework of human art history. Through analysis, we discovered that the qualities of our model’s generated paintings appear to align closely with the characteristics of medieval art in both Eastern and Western traditions. Notably, these two independent artistic heritages exhibit evolution in their styles and techniques during nearly the same historical periods. Therefore, we boldly propose a hypothesis: \textbf{the painting abilities of our model exhibit styles and techniques reminiscent of classical Chinese art from the Wei and Jin Dynasties as well as Western medieval traditions}. 

Usually, quantitative evaluations are well-suited for assessing highly mature technologies, where specific performance of the model or functionalities of the system can be rigorously measured. However, for emergent capabilities like those exhibited by our model—developed without training data and within a relatively short inference time—conventional evaluation metrics may fail to capture the full picture of our model's potential. We suppose that a relatively subjective evaluation framework, such as the perspective of artistic critique, is uniquely positioned to identify and evaluate these abilities. Via artistic comparison and evaluation, we offer insights into our model’s resonance with historical artistic practices, providing a richer understanding of its emergent capabilities. 

In this section, we provide a comprehensive analysis of machine-generated paintings from an artistic perspective. (1) For line drawings shown in \( S_{\text{black}}\),  we examine the model's ability in handling spatial perspective, balancing abstraction and precision, and capturing scenario complexities in its depicted scenes. (2) For colouring shown in \( S_{\text{colour}}\), we explore the richness and harmony of the colour palette, assessing how these qualities echoed medieval art from both stylistic and technical perspectives. (3) Lastly, we delve into two painting techniques showcased in our generated paintings: the five-value system as well as hatching. These techniques reveal a sophisticated handling of light, shadow, and dynamics, underscoring the model's  human-like painting ability with recognition accuracy. Together, these analyses not only highlight human-like painting capabilities of our approach but also suggest the potential to explore a deep connection between the emergent abilities of AI and the evolutionary patterns of human painting. 


\begin{figure*}
    \includegraphics[width=1\linewidth]{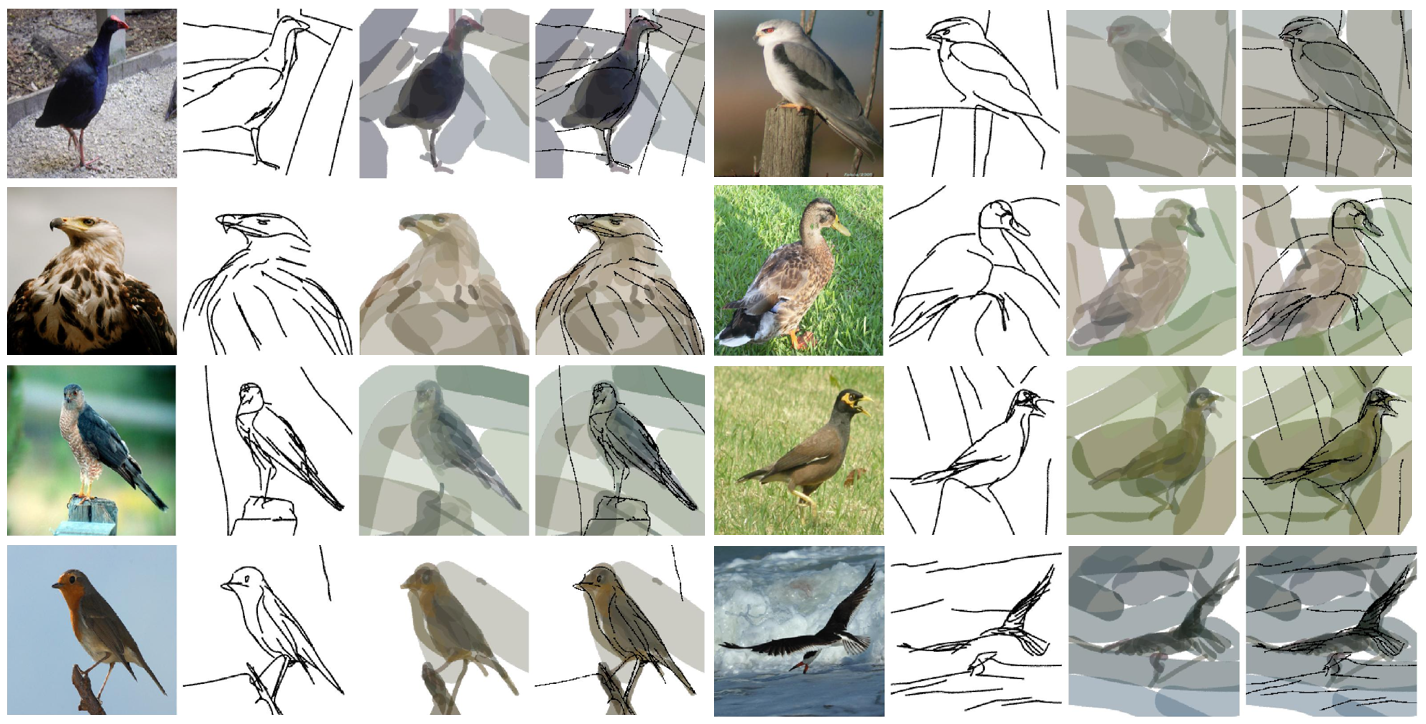}
    \vspace{-4mm}
    \caption{Paintings of Bird. These generated paintings demonstrate how the line drawing technique unique to ancient Chinese art is showcased by our model.}
    \label{fig:more_birds}
\end{figure*}

\begin{figure*}
    \centering
    \includegraphics[width=1\linewidth]{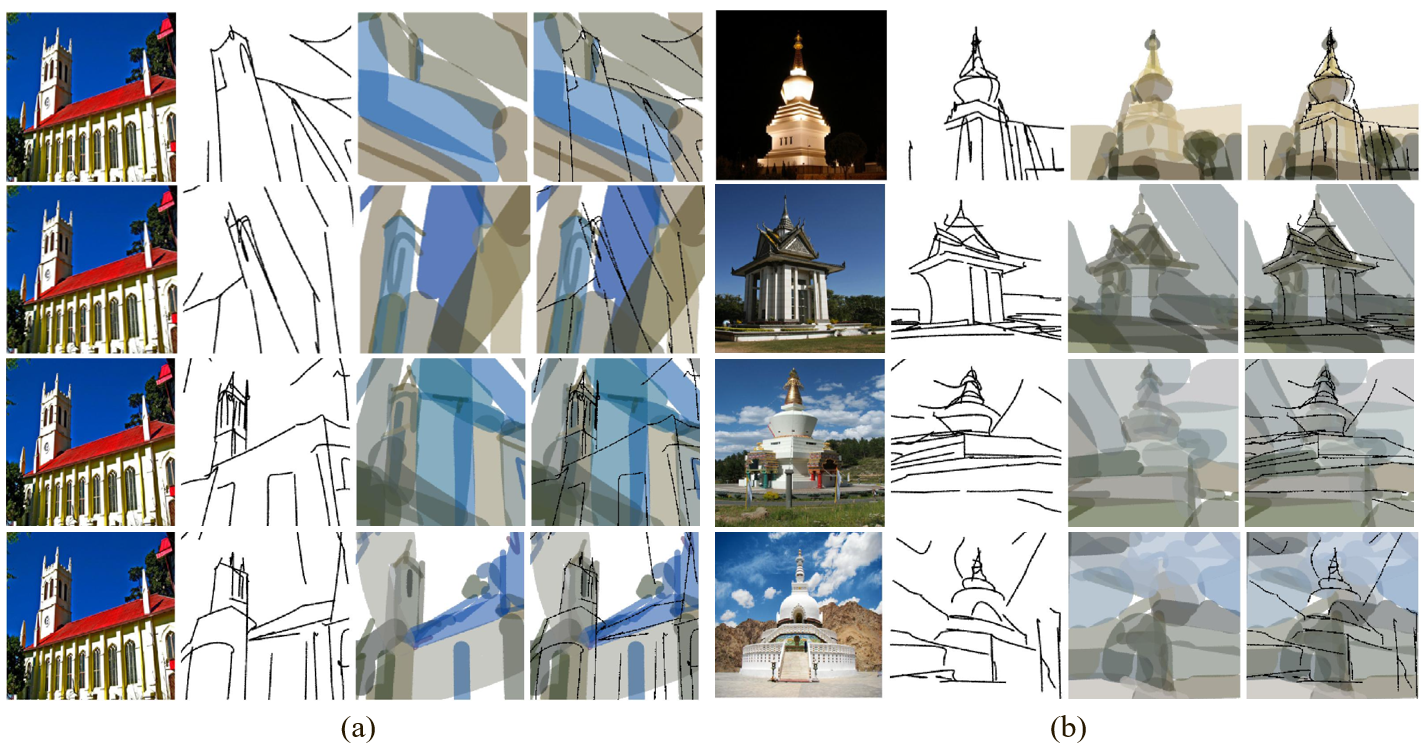}
    \caption{Paintings of Church and Stupa. These generated paintings demonstrate how the flattened and two-dimensional drawing style common in Western medieval art is showcased by our model.}
    \label{fig:church_and_stupa}
    \vspace{-4mm}
\end{figure*}

\subsection*{Line Drawings}
As shown in Fig.~\ref{fig:more_birds}, our model demonstrates a type of line-drawing technique, known in traditional Chinese art as \textit{xiànmiáo}, which serves as a fundamental method of shaping and modeling in Chinese painting. By leveraging variations in line weight, thickness, and other brushwork characteristics, \textit{xiànmiáo} achieves a concise and precise depiction of various subjects. Representative examples include the \textit{tiě xiànmiáo} by Gu Kaizhi of the Eastern Jin Dynasty (c.348–c.409 CE) (Fig.~\ref{fig:linedrawing}(a)) and the \textit{Chuncai-tiao} by Wu Daozi of the Tang Dynasty (c.680–c.759 CE) (Fig.~\ref{fig:linedrawing}(b)). However, our model's demonstration of this line-drawing style lacks a nuanced understanding of spatial depth, resulting in a focus on outlining simple structures of birds. On the other hand, the comprehension of spatial perspective only began to emerge in Chinese art during the Song Dynasty, when painters like Li Gonglin (c.1049-c.1106 CE) demonstrated a sophisticated grasp of spatial relationships and depth in their works (Fig.~\ref{fig:linedrawing}(c)). 

\begin{figure*}
    \centering
    \includegraphics[width=1\linewidth]{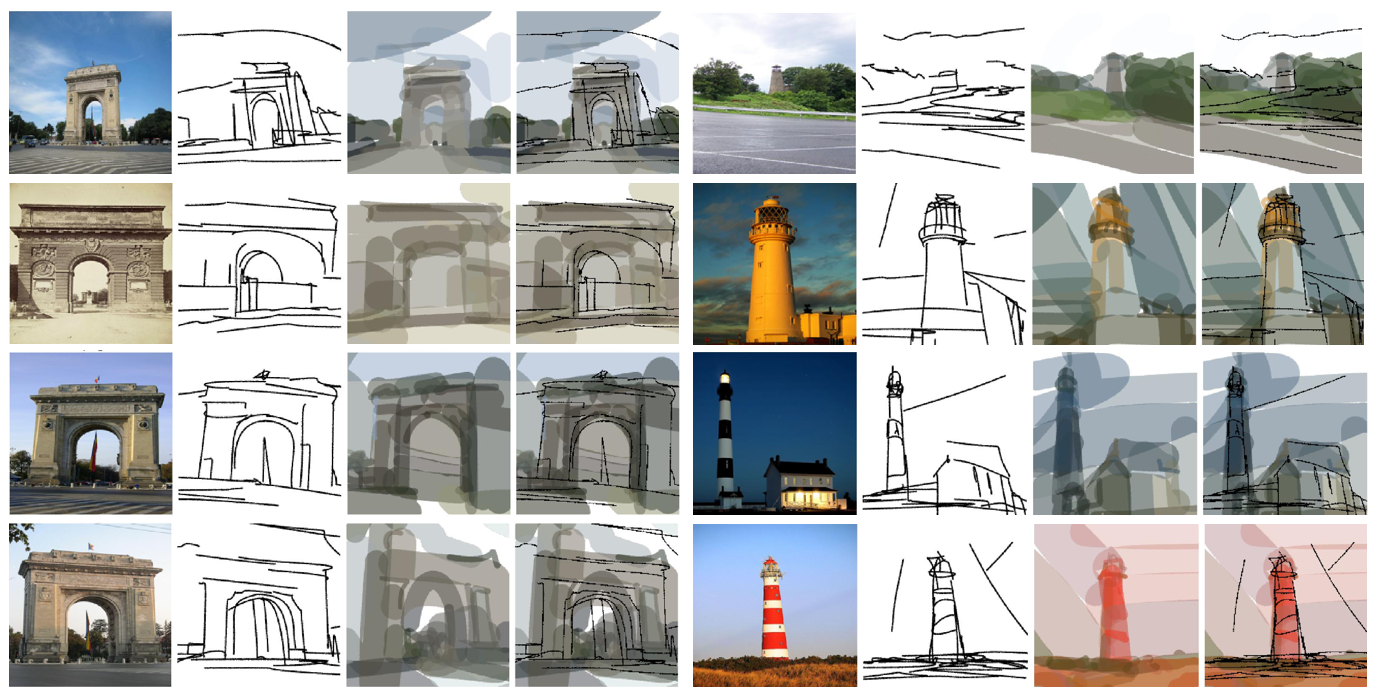}
    \vspace{-4mm}
    \caption{More painting of architecture. Examples are selected from the "triumphal arch" and "beacon" categories.}
    \label{fig:more_arc}
    \vspace{-4mm}
\end{figure*}

This lack of spatial understanding is also evident in the depiction of architectural structures such as churches. In Fig.~\ref{fig:church_and_stupa}(a), the machine-generated paintings exhibit an exceptional emulation of Western medieval art (Fig.~\ref{fig:medieval}), particularly in their flattened, two-dimensional appearance. The paintings generated with 16 and 32 strokes lack the characteristics of linear perspective (or point-projection perspective), as observed in the frontal depiction of the church’s main tower. This reflects the simplified compositional features of medieval art, as artists of that era had yet to develop a mature understanding of perspective. Moreover, the exaggerated prominence of the church’s main tower aligns with the disproportional characteristics typical of that period, where elements at varying distances were often represented at the same scale. In Fig.~\ref{fig:church_and_stupa}(b), the images of the stupa are captured from a nearly frontal perspective, emphasizing their symmetrical and geometric structure. Compared to the complexities of handling the church’s perspective in Fig.~\ref{fig:church_and_stupa}(a), the frontal view of the stupa provides the model with input images that exhibit minimal distortion. The clear vertical and horizontal lines, as well as the distinct geometric shapes of domes, spires, and rectangular bases, enable the machine to more easily understand and render the architectural structure. These machine-generated paintings not only accurately capture the stupa’s symmetrical geometric form but also enhance the depiction with layered colour blocks, introducing depth and dimensionality to the image.

From the examples of bird, church, and stupa, we claim that our model shows line-drawing techniques similar to ancient Chinese art, yet its depiction of three-dimensional structures remains limited, echoing the lack of perspective evident in medieval church paintings in the West. \textbf{These stylistic features indicate that our method not only reflects the common themes of medieval art in scene selection, particularly religious architecture like church, but also naturally reproduces the artistic techniques of the medieval period. These include the distinctive line-drawing methods of Eastern art as well as the flattened stylistic approach of Western medieval church paintings.}

Fig.~\ref{fig:more_arc} showcases additional examples of architectural paintings, such as those of the "triumphal arch" and the "beacon." For instance, in the third row on the left, the machine-generated painting effectively captures the structural integrity and symmetry of the arch with precise contour lines, evoking the style of medieval architectural drawings. Key features of the arch, including its columns, decorative elements, and curved openings, are rendered in a highly abstract yet accurate manner. The addition of tonal layers further enhances the sense of depth and volume. This validates the role of our scene complexity evaluation system in filtering painting scenarios that reflect specific artistic periods. Furthermore, our method reflects the machine's inherent ability to perceive and represent the world in a manner closely aligned with historical artistic practices. Just as medieval art was constrained by the cultural and technological limitations of its time, our method demonstrates emergent painting abilities within restricted scene selection, limited abstraction, and constrained stroke and colour ranges, producing results that are strikingly consistent with the artistic characteristics of the medieval period.

\begin{figure*}
    \centering
    \includegraphics[width=1\linewidth]{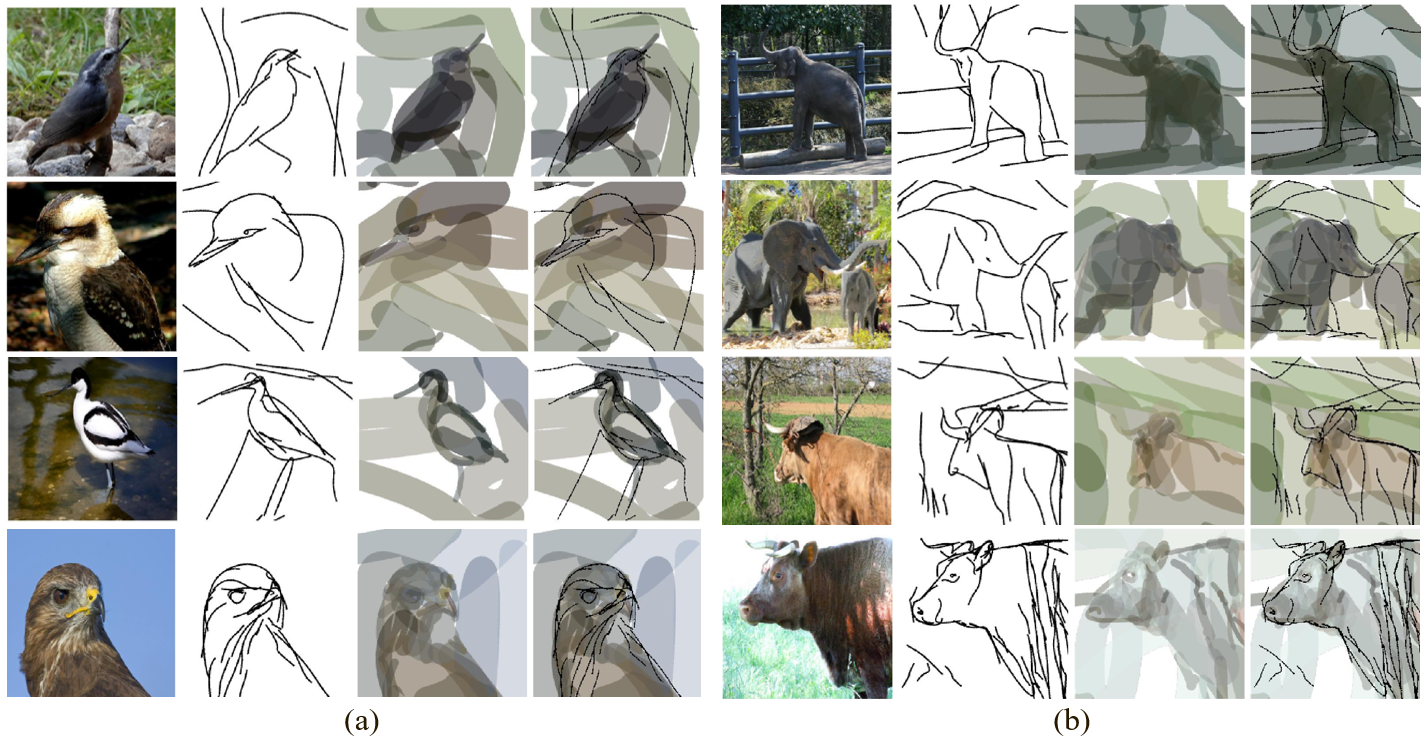}
    \vspace{-4mm}
    \caption{Paintings of Birds, Elephants, and Cows.}
    \label{fig:bird_and_cow}
\end{figure*}

\subsection*{Colouring}

When discussing the consideration of painting scenarios, it is worth noting the performance of our method in depicting subjects particularly birds and cows.

\begin{figure}
    \centering
    \includegraphics[width=1\linewidth]{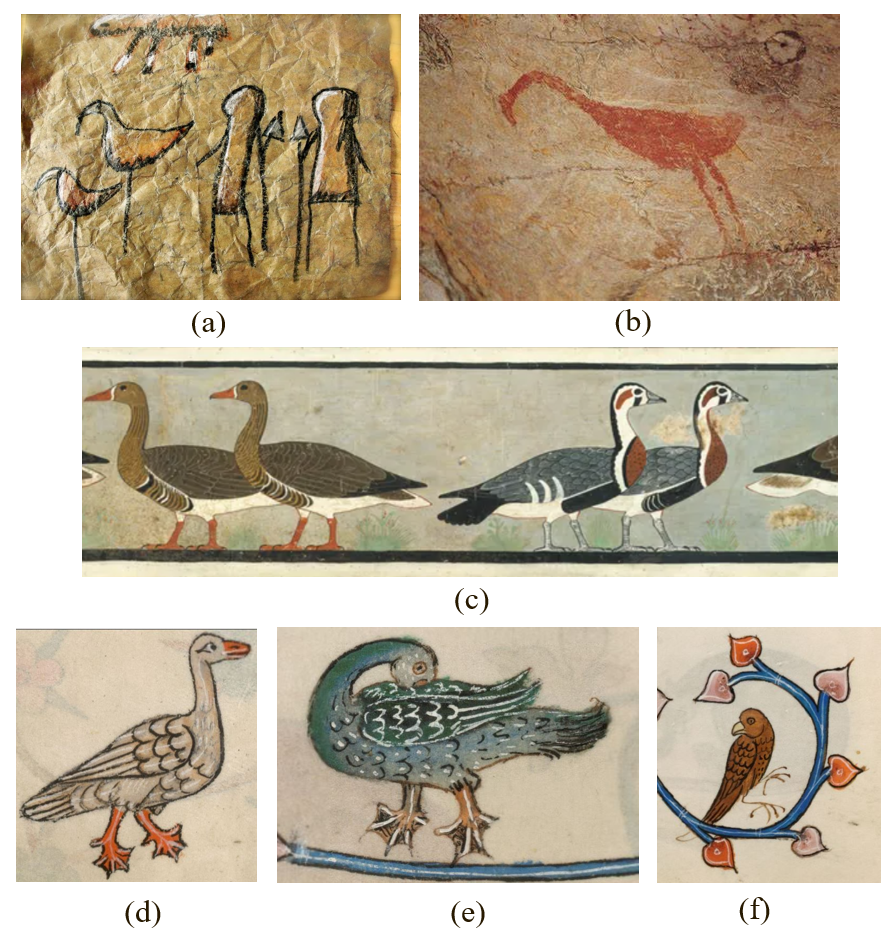}
    \vspace{-4mm}
    \caption{Depiction of birds in Western Art. (a) Birds in Lascaux cave paintings. (b) An aboriginal cave painting from Australia. (c) Tomb painting \textit{Meidum Geese} in ancient Egypt. (d)-(f) Birds details from The Luttrell Psalter.}
    \label{fig:cave_bird_west}
    \vspace{-3mm}
\end{figure}

\begin{figure}
    \centering
    \includegraphics[width=1\linewidth]{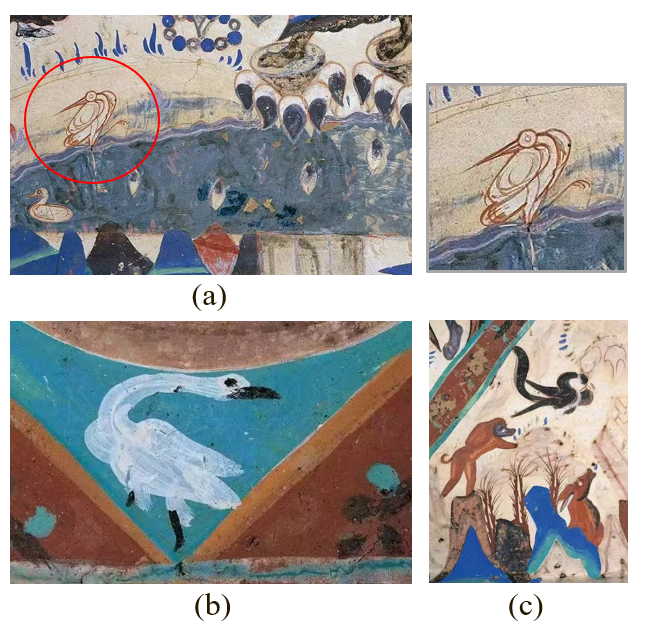}
    \vspace{-4mm}
    \caption{Mogao Cave Paintings. (a) \textit{Waterfowl}, Cave 285, Western Wei Dynasty. (b) \textit{White Goose in Lotus Pond}, Cave 435, Northern Wei Dynasty. (c) \textit{Baboon and Gentoo}, Cave 249, Western Wei Dynasty. (a) shows a line drawing of the bird before coloring, while (b)-(c) demonstrate the technique of overlaying color onto the line drawing.}
    \label{fig:cave_bird_east}
    \vspace{-3mm}
\end{figure}

\begin{figure}
    \centering
    \includegraphics[width=1\linewidth]{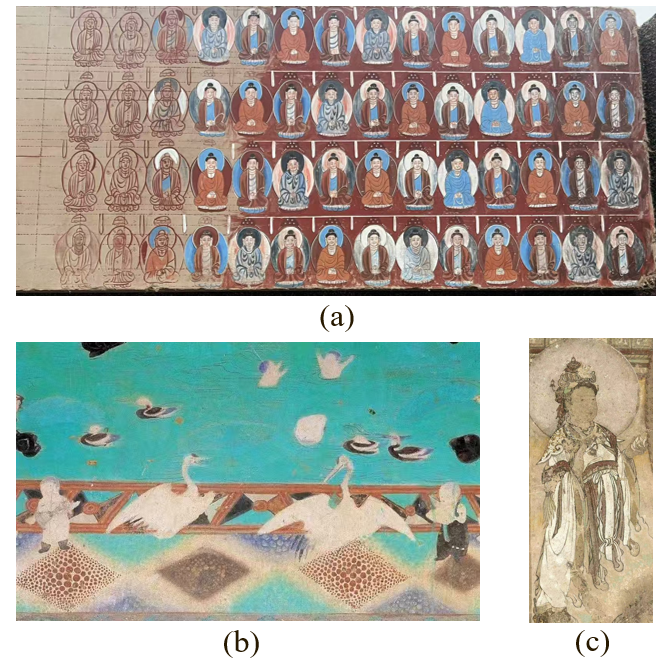}
    \vspace{-4mm}
    \caption{More Mogao Cave Paintings. (a) The process of Dunhuang frescoes involves line drawing first, followed by filling in colors within the outlined areas. (b) \textit{Two Cranes Dancing}, Cave 148, Tang Dynasty. (c) \textit{Qian-Shou-Qian-Yan Guanyin} Painting, Yuan Dynasty).}
    \label{fig:cave_bird_east_2}
    \vspace{-3mm}
\end{figure}

Birds, as an enduring theme throughout art history, have been widely depicted across various styles and eras due to their simplicity and symbolic significance. As shown in Fig.~\ref{fig:cave_bird_west}, the earliest representations of birds can be traced back to Paleolithic cave paintings, such as those in the Lascaux caves in France, where birds were depicted alongside other animals. These early artworks primarily reflect an observation of nature and symbolic expression, with birds often included as part of the imagery to convey hunting targets. In ancient Egyptian art, birds appeared as symbolic figures. By the Middle Ages and early Renaissance, the depiction of birds in art had become more focused on realism and symbolism, particularly in natural history and still-life paintings. For instance, in the medieval \textit{Luttrell Psalter}, birds were rendered with meticulous brushstrokes to capture their posture and feather details. In Eastern art, the depiction of birds underwent a series of stylistic evolutions. Notably, during the Northern Wei Dynasty, birds were primarily illustrated using line drawings overlaid with colour blocks (see Fig.~\ref{fig:cave_bird_east}). By the Tang Dynasty, these colour blocks were more rigorously confined within the contours of the lines, a technique evident in the Dunhuang murals in Mogao Caves and the Yuan Dynasty's \textit{Qian-Shou-Qian-Yan} painting (see Fig.~\ref{fig:cave_bird_east_2}).

As shown in Fig.~\ref{fig:bird_and_cow}(a), our machine-generated paintings capture the essential features of birds accurately, such as their form and texture, achieving at least the artistic level of medieval depictions of birds. The precise contour lines emphasize birds' shapes, while the use of layered colours adds depth and texture to the paintings. Notably, the machine's colouring does not strictly adhere to the contours, a method that closely resembles the overlaying technique observed in artworks in Northern Wei Dynasty. Moreover, these paintings exhibit a progression toward the filling-style colouring of the Tang Dynasty, where colours were confined more rigorously to the outlines. This transition, supported by extensive archaeological evidence, suggests that the colour-filling or colour-blocking technique Thus, we claim that the emergent painting ability of our model aligns with a style that bridges the period between the Northern Wei and Tang to Yuan Dynasties.

\begin{figure}
    \centering
    \includegraphics[width=1\linewidth]{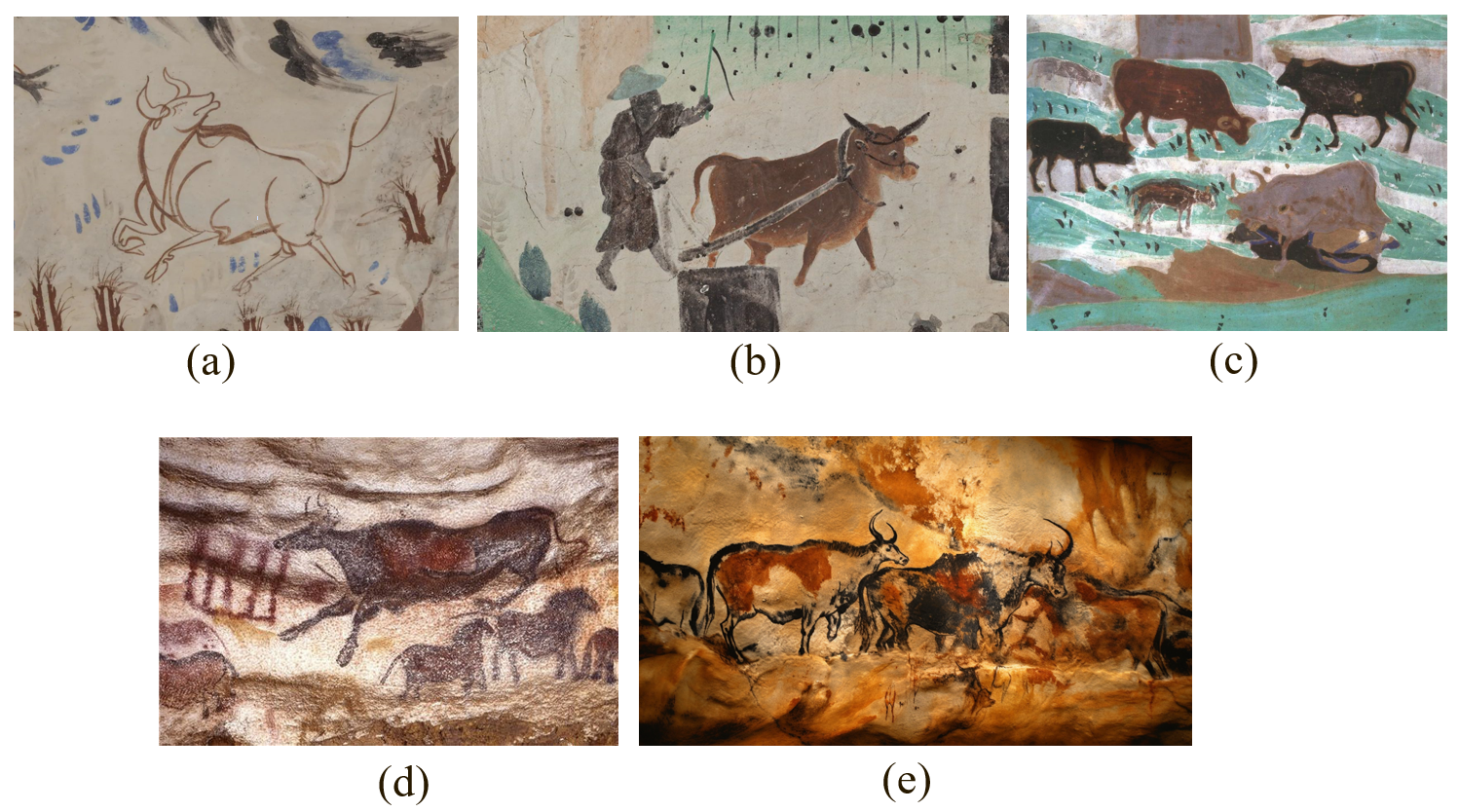}
    \caption{Depiction of cows in Cave Paintings. (a) \textit{Frightened Bison}, found in Cave 249, Mogao Caves. (b) \textit{Farming Map}, found in Cave 23, Mogao Caves. (c) \textit{The Cow King and the Herd of Cows}, found in Cave 23, Mogao Caves. (d)-(e) Lascaux cave paintings of cows.}
    \label{fig:cave_cow}
    \vspace{-5mm}
\end{figure}

Besides birds, paintings of other animals also exhibit styles closely aligned with medieval art. As shown in the first two rows of Fig.~\ref{fig:bird_and_cow}(b), the distinct contour lines and the intricate curvature of the elephant's trunk are rendered with precision. Notably, in the last two rows of Fig.~\ref{fig:bird_and_cow}(b), the depiction of the cow’s form is strikingly accurate. Since the advent of cave painting, cows have served as highly representative and universal symbols, often used to convey essential survival information, such as the use of hunting tools. Similar to the styles observed in the Lascaux and Altamira cave paintings (see Fig.~\ref{fig:cave_cow}), the bold contour lines in these paintings capture the fundamental form and posture of the cow, emphasizing key features such as its horns and body curvature. Beyond their prehistoric significance, cows also played an important symbolic role in medieval art, frequently appearing in tapestries, religious iconography, and decorative works.

Animals like birds, cows, and horses have long been common and representative subjects in art, appearing extensively in works predating the Middle Ages in both Western and Eastern art traditions. This prevalence highlights humanity's mastery of animal depiction even before the medieval period. \textbf{Similarly, the paintings generated by our model not only capture the intricate details of birds and cows through precise linework but also employ layers of colour blocks to convey depth and texture, showcasing a painting ability consistent with medieval artistic practices}. More specifically, our model demonstrates artistic techniques closely aligned with the Northern Wei Dynasty. These features attest to the model’s ability in painting animals like birds and cows, validating that its emergent painting ability is at least on par with the level of medieval art. 

\subsection*{Painting Technique: the Five-Value System}    

Our machine-generated paintings emulate human techniques for depicting light and shadow with good recognizability, skillfully applying the fundamental principles of the \textit{Five-Value System}. The Five-Value System divides light and shadow into five zones: highlights, form shadows, core shadows, cast shadows, and reflected light. In the paintings of still-life such as apple, the highlight on the right side of the apple is accurately rendered, achieved through the use of blank spaces left by the brush. In the darker areas facing away from the light source, the application of form shadows captures a sense of depth and volume. It is evident that our model has developed an awareness of the Five-Value System in its paintings, particularly in still-life depictions. However, this technique only emerges under specific conditions. When the subject itself has a strong sense of three-dimensionality, the machine produces paintings with pronounced depth; when the subject appears flat, the resulting painting reflects that flatness. Interestingly, we found it natural to analogize still-life painting to an art exam, making it relatively straightforward to evaluate these works quantitatively from a grading perspective. According to professional artists, the paintings of apples and oranges exhibit accurate colors, with core shadows and highlights rendered precisely. Based on this assessment, we can further grade the results quantitatively: \textbf{this mastery of the Five-Value System enables our model to outperform approximately 30\% of untrained individuals after two weeks of painting practice}. See the supplementary material for more examples.

\subsection*{Painting Technique: Hatching}

In the paintings of water, the deliberate use of parallel lines effectively showcases our model’s ability to convey the texture of water, the interplay of light and shadow, and the dynamic flow of movement. This technique reflects a refined application of the \textit{hatching} technique, a method in Western art dating back to the Middle Ages and the Renaissance. Masters such as \textit{Leonardo da Vinci} and \textit{Albrecht Dürer} are renowned for their exquisite use of parallel, wavy, and cross-hatched lines in their black-and-white sketches and engravings, particularly in their depictions of ripples, waves, and reflections on water. For instance, da Vinci’s Deluge with Stormy Skies employs dense lines to capture the texture and movement of water surfaces precisely. This technique is also reflected in the paintings generated by our model. The details of ripples are rendered through a series of parallel lines, with subtle directional variations conveying the dynamic nature of water. The reflections on the water surface are expressed through variations in line spacing, effectively illustrating differences in brightness. This nuanced treatment of lines not only enhances the depth of the composition but also captures the layers of flowing water, echoing the classical approaches of early Renaissance artists. However, from the perspective of Eastern art, the model’s mastery of hatching does not yet achieve the “distant water without ripples” principle, where artists in the Tang and Song Dynasty used line density to depict the sparseness and density of water.

By naturally exhibiting this technique in its generated paintings, our model resonates with the artistic practices of the Middle Ages and early Renaissance in terms of scene selection, detail rendering, and artistic expression. The complexity and technical refinement evident in these paintings not only reflect the core characteristics of human artistic styles but also validate the model’s significant performance in aligning with historical artistic traditions. See the supplementary material for more examples.


\section{Conclusion}

In this paper, we propose a novel approach to explore whether human-like painting ability can emerge through recognition-driven evolution, focusing on scenario complexity, recognition accuracy, and abstraction with colour richness. Our painting model combines a stroke branch and a palette branch to generate human-like paintings, learning an optimal colour palette for strokes. We quantify visual communication efficiency through a high-level recognition module, minimizing the semantic embedding distance between the generated painting and the input image. The scenario complexity estimator addresses the gap in previous methods by estimating scenario complexity for painting generation. Experimental results show our model achieves high recognition accuracy with minimal strokes and colours, while colourized strokes enhance visual appeal and information conveyance. Our method also demonstrates potential as an efficient image compression technique, outperforming traditional methods. These findings highlight our approach's contribution to understanding the evolution of painting and its applications in image compression and artistic content generation.

{
    \small
    \bibliographystyle{ieeenat_fullname}
    \bibliography{main}
}

\end{document}